\pdfoutput=1

\documentclass[11pt]{article}

\usepackage[preprint]{acl}

\usepackage{inconsolata}
\usepackage{times}
 
\usepackage{booktabs}
\usepackage{multirow}
\usepackage{tablefootnote}
\usepackage{array}
\usepackage{pdfpages}
\usepackage{colortbl}

\usepackage{amsmath}
\usepackage{amsfonts}
\usepackage{pifont}
\usepackage{makecell}
\usepackage{longtable}
\usepackage{geometry}

\usepackage{graphicx}
\usepackage{float}
\usepackage{subcaption}
\usepackage{paralist}
\usepackage{enumitem}
\usepackage{titlesec}

\usepackage{xcolor}

\usepackage[T1]{fontenc}

\usepackage{amssymb}

\usepackage[utf8]{inputenc}

\usepackage{microtype}

\usepackage{inconsolata}

\usepackage{graphicx}

\title{GenTool: Enhancing Tool Generalization in Language Models through Zero-to-One and Weak-to-Strong Simulation}

\author{Jie He$^1$\thanks{Work was done at Microsoft.}\footnotemark[2], Jennifer Neville$^2$\thanks{Corresponding author: j.he@ed.ac.uk, jenneville@microsoft.com,  pei.zhou@microsoft.com.}, Mengting Wan$^2$, Longqi Yang$^2$, \\ \textbf{Hui Liu$^2$}, \textbf{Xiaofeng Xu$^2$}, \textbf{Xia Song$^2$}, \textbf{Jeff Z. Pan$^1$}, \textbf{Pei Zhou$^2$}\footnotemark[2]\\
 $^1$ School of Informatics, University of Edinburgh, UK \\
 $^2$ Microsoft Corporation \\
 }

\begin{document}
\maketitle
\begin{abstract}
\textbf{L}arge \textbf{L}anguage \textbf{M}odels (LLMs) can enhance their capabilities as AI assistants by integrating external tools, allowing them to access a wider range of information. While recent LLMs are typically fine-tuned with tool usage examples during supervised fine-tuning (SFT), questions remain about their ability to develop robust tool-usage skills and can effectively generalize to unseen queries and tools. In this work, we present \textbf{GenTool}, a novel training framework that prepares LLMs for diverse generalization challenges in tool utilization. Our approach addresses two fundamental dimensions critical for real-world applications: \textit{Zero-to-One Generalization}, enabling the model to address queries initially lacking a suitable tool by adopting and utilizing one when it becomes available, and \textit{Weak-to-Strong Generalization}, allowing models to leverage enhanced versions of existing tools to solve queries. To achieve this, we develop synthetic training data simulating these two dimensions of tool usage and introduce a two-stage fine-tuning approach: optimizing tool ranking, then refining tool selection. Through extensive experiments across four generalization scenarios, we demonstrate that our method significantly enhances the tool-usage capabilities of LLMs ranging from 1B to 8B parameters, achieving performance that surpasses GPT-4o. Furthermore, our analysis also provides valuable insights into the challenges LLMs encounter in tool generalization.


\end{abstract}
\section{Introduction}
Tool learning has emerged as a crucial capability for \textbf{L}arge \textbf{L}anguage \textbf{M}odels (LLMs), enabling them to expand their functionality through external tool integration \cite{huang-etal-2024-planning, tang2023toolalpaca, patil2023gorilla}. By interfacing with external tools, LLMs can dynamically access real-time information, validate responses against external knowledge bases, and improve outputs through iterative feedback \cite{deng2023mind2web, wang-etal-2024-llms-imaginarium}. This capability fundamentally enhances LLMs' ability to process and respond to real-world information beyond their pre-trained knowledge.

\begin{figure}[!tpb]
    \centering
    \includegraphics[width=0.9\linewidth]{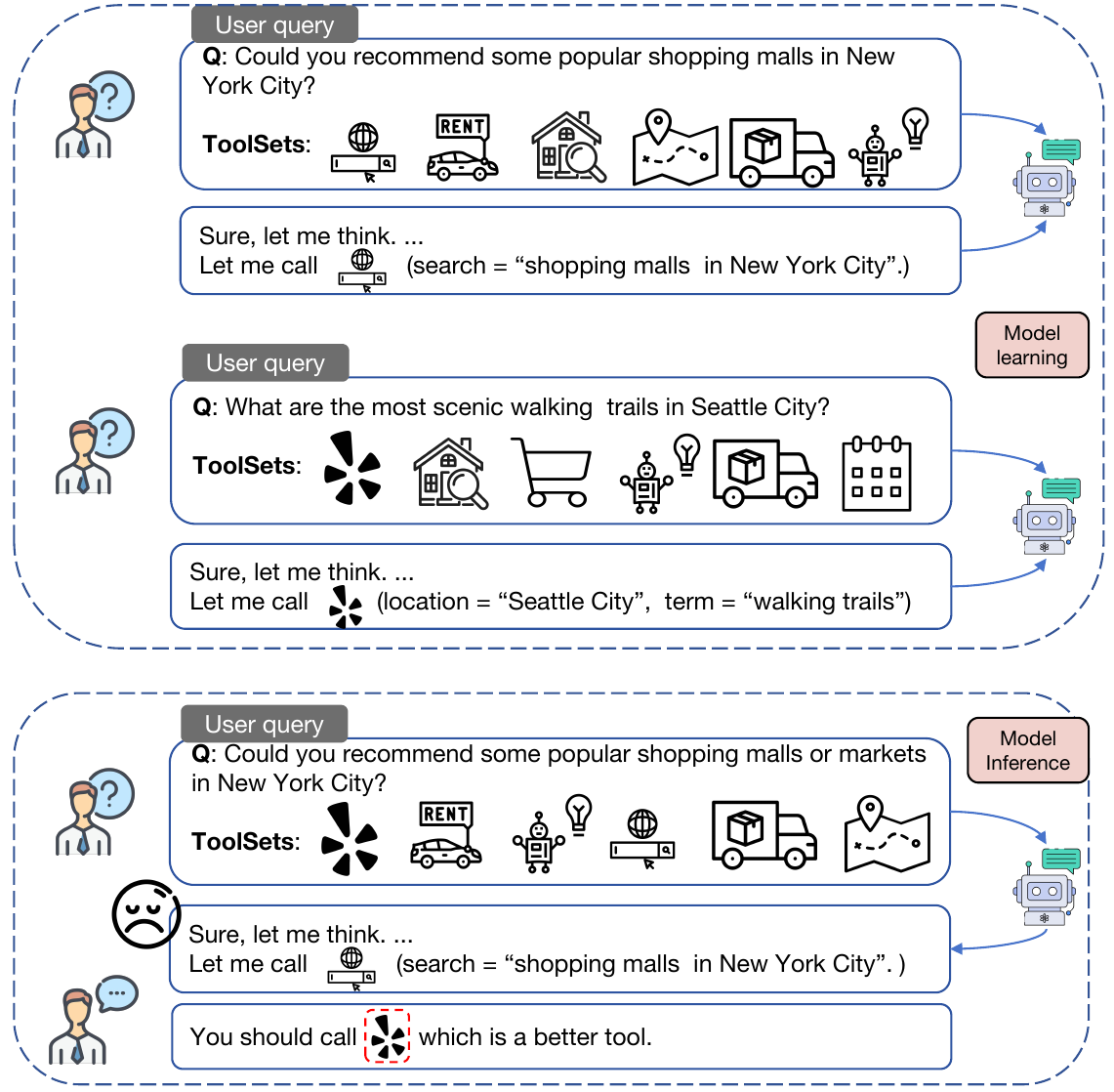}
    \caption{ 
    An example illustrating tool generalization challenges in selecting the most suitable tool for a user query. While the model was trained on tools like Yelp and Web-Search, and encountered the same query:  ``Could you ... in New York City?'', it struggles during testing to select the more appropriate Yelp tool over Web-Search for the same query during test. 
    }
    \label{fig:intro}
    \vspace{-0.6cm}
\end{figure}

The current paradigm for tool utilization follows a standard workflow \cite{qu2024toollearninglargelanguage}: given a user query and available tool documentation, LLMs identify appropriate tool, extract required parameters, obtain information from the tool and synthesize final outputs. Two primary approaches have been developed to enable this capability: in-context learning, which leverages tool documentation and examples within the model's context \cite{hsieh2023tooldocumentationenableszeroshot}, and fine-tuning on collected tool usage data \cite{qin2024toolllm, mekala-etal-2024-toolverifier}. However, both approaches face fundamental limitations. In-context learning is constrained by context length restrictions, preventing comprehensive tool understanding \cite{DBLP:journals/corr/abs-2303-17491, paranjape2023artautomaticmultistepreasoning}. Fine-tuning methods risk over-fitting to specific tools and usage patterns, compromising generalization to unseen tools and queries. These limitations pose significant challenges for real-world applications, where new tool categories and usage patterns constantly arise as shown in Figure \ref{fig:intro}.

To tackle these challenges, we introduce \textbf{GenTool}, a novel framework designed to enhance LLM's tool generalization capabilities across two core dimensions: \textbf{zero-to-one generalization}, where a query initially has no available useful tool, and \textbf{weak-to-strong generalization}, where models must adapt from a weak tool which is ineffective in fully addressing the query to a strong one which can help provide an answer that perfectly matches the query. Through these dimensions, we identify four evaluation scenarios: \textit{seen\_query\_unseen\_tool}, \textit{seen\_query\_seen\_tool}, \textit{unseen\_query\_unseen\_tool}, and \textit{unseen\_query\_seen\_tool}.

To enable effective training across the two important dimensions, we developed a high-quality synthetic training dataset. This dataset comprises 834 new synthetic tools and their descriptions. For each tool, we generated 10 diverse queries, resulting in 8,515 distinct queries and 33,286 high-quality training samples. Each sample includes structured query-tool pairs and detailed parameter information. Additionally, we complement our data augmentation strategy with a novel two-stage fine-tuning strategy that
explicitly teaches models to rank tools by capability before selection, enabling systematic understanding of tool relationships rather than simple query-tool mappings.


Our extensive evaluation demonstrates GenTool's effectiveness across various model architectures including LLaMA 3.2 1B, LLaMA 3.1 8B, Mistral 7B, and Phi 3B. GenTool achieves state-of-the-art performance across all metrics, significantly outperforming both tuning-free and tuning-based baselines, including GPT-4o, with a 14.28\% improvement in tool selection accuracy. Moreover, GenTool exhibits exceptional generalization capabilities across our four evaluation scenarios. Additionally, ablation studies validate the importance of our ranking mechanism, where its removal leads to performance degradation of up to 10.89\%.

\textbf{Our main contributions are summarized as follows:}
\begin{compactenum}     
\item We propose GenTool, a novel framework that enhances LLMs' generalization capabilities on unseen tools through structured simulation of two critical dimensions. GenTool achieves state-of-the-art performance across all metrics, significantly outperforming various strong baselines.

\item We introduce a two-stage fine-tuning strategy combining tool selection with ranking supervision, enabling enabling robust tool generalization through better understanding of functional differences between similar tools.

    

\item We provide comprehensive empirical analysis revealing key insights into tool generalization, including the limitations of data scaling and the importance of diverse training settings, evidenced by a 67\% performance drop when training exclusively on test-set-similar cases.

\end{compactenum} 
\section{Problem Definition: Generalization in Tool Learning}
\label{sec2}
Our framework aims to improve LLM generalization across diverse query-tool combinations. This section defines the tool generalization problem and presents a systematic framework for its evaluation.
\subsection{Preliminaries}

To analyze tool learning, we first define the core components of our framework. Each instance $d$ in our dataset is represented as a tuple $(T_s, q, y, t)$, where the toolset $T_s$ consists of $k$ distinct tools,   $q$   the query submitted to the model,  $t\in T_s$ the ground truth tool  required to resolve the query, and $y$ the tool-calling information  specifing the tool name, parameter names, and their corresponding parameter values.

Each tool $t_i$ within $T_s$ is equipped by metadata that includes its name, descriptions of its parameters, names of expected results, and descriptions of those results. This formalization provides a foundation for assessing the model's ability to generalize across varying combinations of queries and tools, establishing the groundwork for our framework.

\begin{figure*}[!t]
    \centering
    \includegraphics[width=0.8\linewidth]{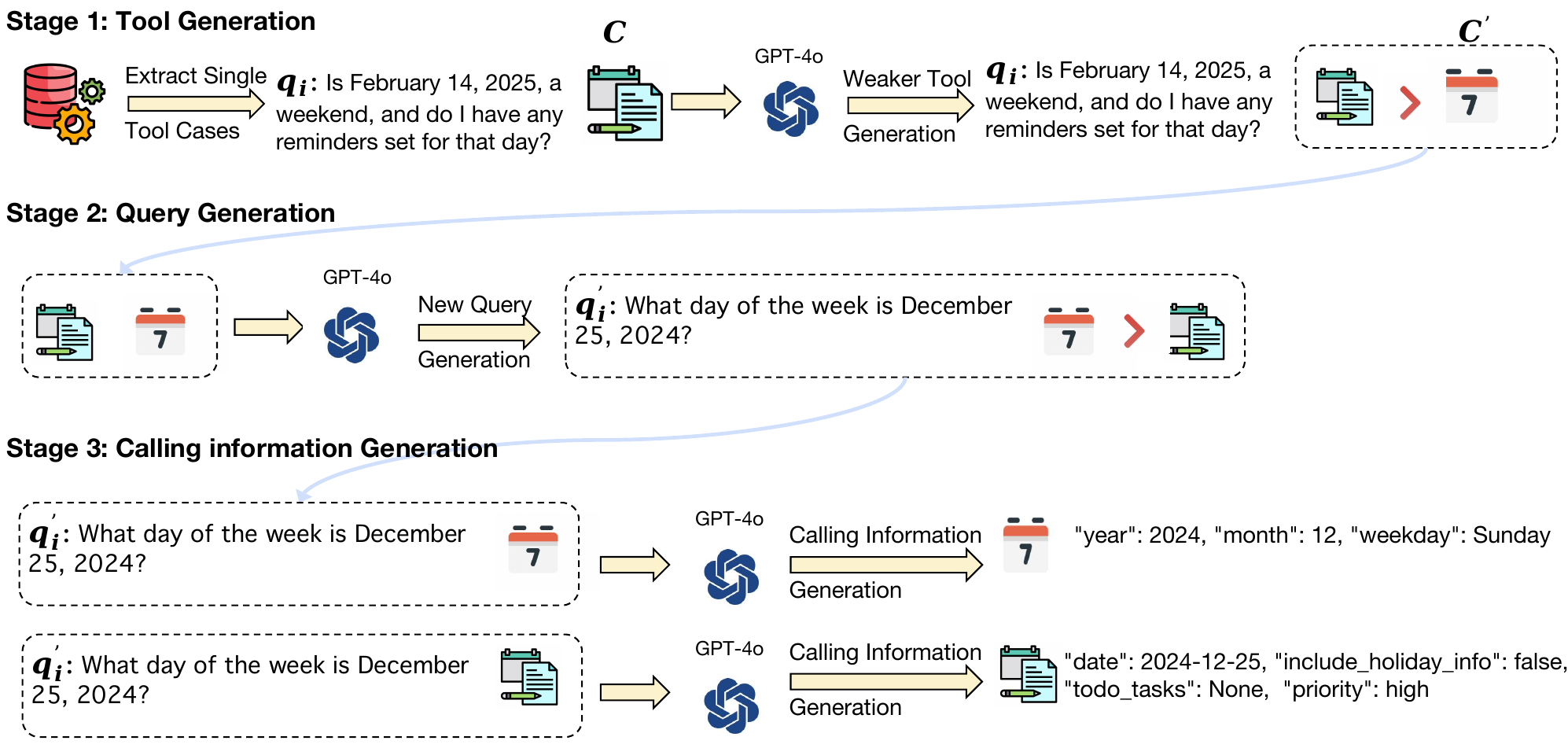}
    \caption{The construction process for synthetic data to simulate the generalization process involves three steps: First, existing datasets are utilized to create new tools. Next, diverse instructions guide the generation of matching queries for these new tools. Finally, corresponding invocation details are created for various tool-query combinations.}
    \label{fig:data_cons}
    \vspace{-0.3cm}
\end{figure*}
\subsection{Four Generalization Scenarios}
\label{sec2.2}
To systematically evaluate a model's generalization capabilities, we define four scenarios based on the familiarity of queries and tools during training.

\paragraph{Seen Query and Unseen Tool}
The training set contains a sample $d_{\text{train}}^1 = (T^1_s, q_1, t_1, y_1^1)$, while the test set includes a sample $d_{\text{test}}^1 =  (T^2_s, q_1, t_{2}, y_1^2)$. The query $q_1$ is identical across training and test sets, but the ground truth tool $t_{2}$ in the test case is not present in any training toolsets $T_s$, making $q_1$ a \textit{seen query} and $t_2$ an \textit{unseen tool}.


\paragraph{Seen Query and Seen Tool}
Both the query and the ground truth tool are present during training. The training set contains    $d_{\text{train}}^1 = ( T_s^2, q_1, t_2, y_1^2 )$, and  $d_{\text{train}}^2 = ( T_s^1, q_2, t_1, y_2^1 )$ and the test set includes $d_{\text{test}}^1 = ( T_s^1, q_1, t_1, y_1^1 )$. $q_1$ and $t_1$ are \textit{seen}.

\paragraph{Unseen Query and Unseen Tool}
Neither the query nor the tool has been observed during training. The training set includes $(T_s^1, q_1, t_1, y_1^1)$, while the test set contains $(T_s^2, q_2, t_2, y_2^2)$, making $q_2$ and $t_2$ both \textit{unseen}.

\paragraph{Unseen Query and Seen Tool}
The test query is unseen, but the ground truth tool appears in training. The training set includes $(T_s^1, q_1, t_1, y_1^1)$, and the test set has $(T_s^1, q_2, t_1, y_2^1)$. Here, $q_2$ is \textit{unseen}, and $t_1$ is \textit{seen}.

These scenarios provide a structured framework for assessing model generalization across queries and tools, offering valuable insights into robustness and adaptability of tool learning models in real-world applications.


\begin{figure*}[t]
    \centering
    \includegraphics[width=0.8\linewidth]{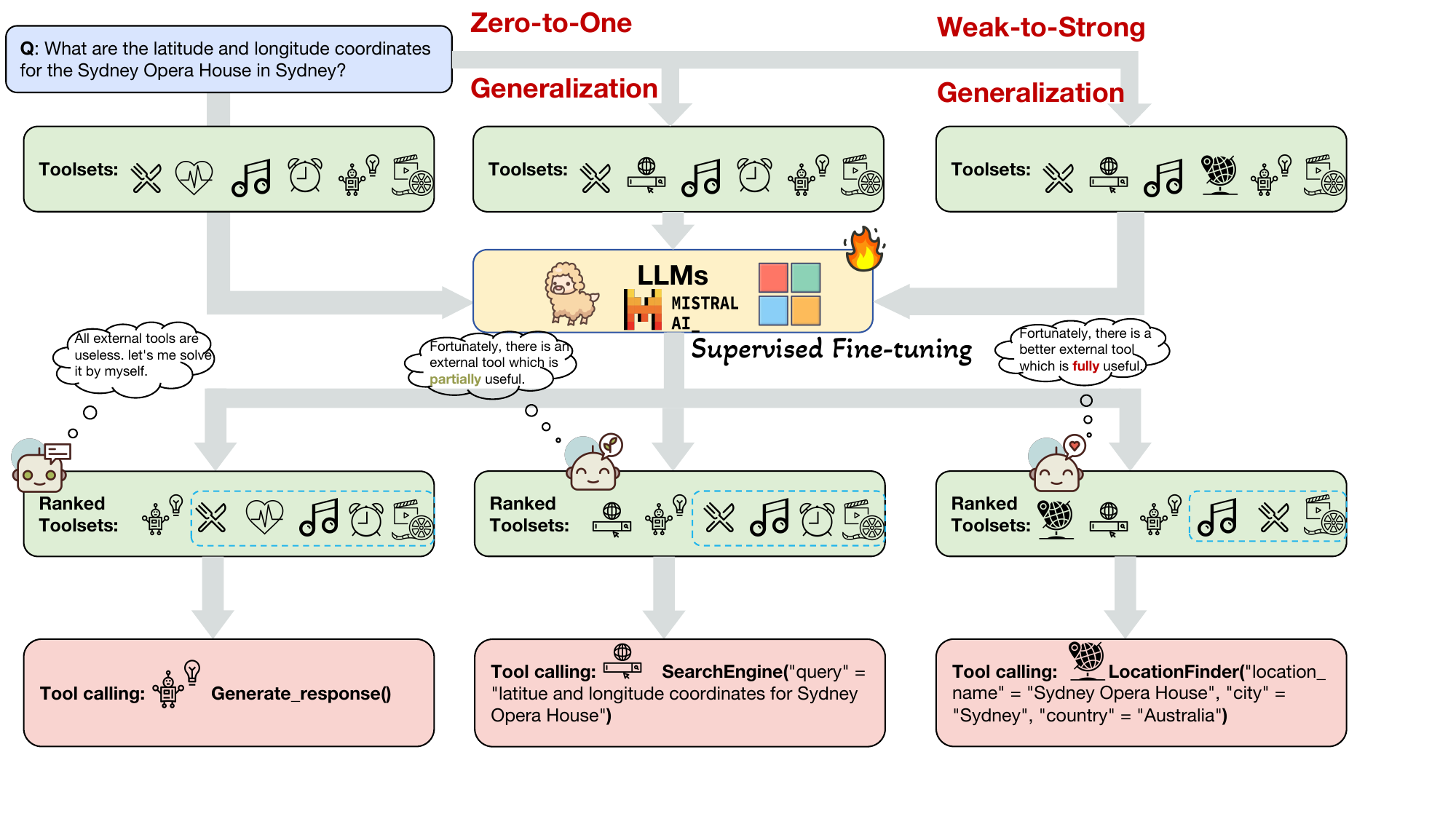}
    \caption{Overview of the GenTool  Framework for Tool Learning and Generalization. Initially, the model handles a query by defaulting to \texttt{generate\_response} when no suitable tool is available. Next, when a relevant tool, \texttt{web\_search}, is added to the toolset, the model selects \texttt{web\_search}, demonstrating zero-to-one generalization training. Later, 
    upon adding \texttt{map\_search}, the model demonstrates weak-to-strong generalization by correctly ranking and selecting it over web\_search and other alternatives.}
    \label{fig:GenTool _framework}
    \vspace{-0.3cm}
\end{figure*}

\section{GenTool : Synthetic Data Generation for Generalization Simulation } 
\label{sec:data_generation}

The gap in generalization remains challenging to address with current data \cite{qin2024toolllm, tang2023toolalpaca}, as models tend to learn biases from the tools available in the training set. To tackle this issue, we propose GenTool, a synthetic data generation framework that simulates generalization scenarios during the training phase.

Our approach is motivated by a key observation: when a model only has access to a basic calendar tool  $\text{C}'$, it defaults to using it. However, when presented with both $\text{C}'$ and a better tool, $\text{C}$, the model should recognize and select $\text{C}$ for complex queries. We employ GPT-4o to generate additional tools that maintain the same format as the original tools in our dataset. For each original tool t, we generate weaker variants that have limited functionality while preserving the API structure.

Figure \ref{fig:data_cons} illustrates our data generation pipeline, which consists of three primary components: tool generation, query generation, and calling information synthesis. Each component is designed to create diverse, high-quality training instances that specifically target generalization capabilities. Details for prompts and data quality checking, please refer to App. \ref{app_datasets}.

\subsection{Tool Generation}
\label{section3.1}


We build upon the UltraTool \cite{huang-etal-2024-planning-creation} dataset, comprising 5,000 high-quality user queries refined by human annotators. For our initial investigation, we focus on the 800 single-tool usage cases (\textbf{original data}). This scope allows us to establish fundamental principles of tool generalization before tackling multi-tool interactions explored in works like MetaTool \cite{huang2024metatool}. 

For each query and its gold tool description, We use GPT-4o to generate two ``weak tools'' which is ineffective in fully addressing the query and fails to provide an answer that perfectly matches the query. For example, for the query  ``What are the best restaurants in New York City, and what cuisines do they serve?'', a weak tool may return the top 10 restaurants but not provide cuisines information, whereas a strong tool could provide both the top 10 restaurants and their cuisines. In this case, the strong tool should be called. This process ensures consistent API formatting while varying tool capabilities.




\subsection{Query Generation}
To capture diverse generalization scenarios, we generate tailored queries for each weak tool. For each tool pair (gold tool $t$ and weak tool $t^{'}$), we prompt GPT-4o to generates 10 queries meeting two criteria: 1) solvable by the weak tool $t^{'}$, and 2) only partially addressable by the gold tool $t$.
This approach forces the model to distinguish tools based on capabilities, and the selection of 10 queries per tool pair provides sufficient variation while maintaining dataset manageability.


\subsection{Calling Information Generation}
Accurate tool invocation is crucial for training. We employ GPT-4o to generate calling information for each query-tool combination. For a given query $q$ and a tool $t$, GPT-4o generates the necessary parameter names and corresponding parameter values required to invoke $t$, ensuring the dataset reflects the realistic process of tool invocation.

\section{GenTool: Training and Evaluation Framework}
This section presents our training and evaluation methodology as shown in Figure \ref{fig:GenTool _framework}. We first introduce two generalization dimensions: \textit{zero-to-one} and \textit{weak-to-strong} generalization. We then describe our test instances construction for evaluating the generalization scenarios outlined in Section\ref{sec2.2}. Finally, we propose a two-stage output mechanism to enhance tool selection accuracy.



\subsection{Training Strategy for Two Generalization Dimensions}
Our training addresses two real-world generalization dimensions: \textit{zero-to-one} and \textit{weak-to-strong}. We first describe our toolset generation process, which is fundamental to both frameworks.


\subsubsection{Toolset Generation}
For each query $q$, the model is provided with a toolset $T_s$ consisting of $k$ potential candidate tools, including the gold tool  $t$ required for the query. The construction process involves: 1) Computing embeddings for all tools in the dataset using the text-embedding-ada-003 model \cite{OpenAI2023EmbeddingModel}; 2) For each $(q, t)$ pair, retrieving the $k-1$ most similar tools based on cosine similarity for $T_\text{gold}$ to generate the $T_s$; 3) Ensuring diversity by excluding redundant tool-query pairs in which  tools already paired with the same query during retrieval. Considering the input length limitation of the model, we set \(k = 5\). To handle cases where no suitable tool exists in the toolset, we introduce \texttt{generate\_response} as an additional tool, which enables the model to generate direct responses when necessary.

\subsubsection{Zero-to-One Generalization Training}
Given the dataset $\text{QT} = \{(q_1, t_1, q_1^{'}, t_1^{'}), \dots, (q_n, t_n, q_n^{'}, T_n^{'})\}_{i=1}^n$, we first sample a subset $\text{QT}_{\text{pure\_tr}} = \{(q_i, t_i, q_i^{'}, t_i^{'})\}_{i=1}^k$ to  construct training instances simulating \emph{zero-to-one} and \emph{weak-to-strong} transitions. The remaining portion of the dataset is defined as $\text{QT}_{\text{test}} = \{(q_j, T_j, q_j^{'}, T_j^{'})\}_{j=k+1}^{n}$ serves as our evaluation set for testing generalization under four distinct scenarios (See \ref{sec:scenario_generation}). 

For each $\{q, t, q^{'}, t^{'}\}_i \in \text{QT}_{\text{pure\_tr}}$, we constructed two training examples:

\begin{compactenum} 
    \item   $d_{\text{train}}^1 = ( T_s^{N}, q_1, \text{None}, y_{N} )$, where \( t = \text{None} \), it indicates that the toolsets do not contain a tool matching the given query. In this case, $y_{N} = \texttt{generate\_response()}$.
    \item $d_{\text{train}}^2 = ( T_s^1, q_1, t_1, y_1^1)$, where the model should select $t_1$.
\end{compactenum} 
By including both $d_{\text{train}}^1$ and $d_{\text{train}}^2$ in the training set, the model learns to distinguish between cases where no tools in the toolset can solve the query and cases where a valid tool exists, thereby simulating \textit{zero-to-one} generalization. 
Similarly, \( q' \) and \( t' \) follow similar construction patterns for \textit{zero-to-one} generalization training example.

\subsubsection{Weak-to-Strong Generalization Training}
As described in Section \ref{sec:data_generation}, for each $\{q, t, q^{'}, t^{'}\}_i \in \text{QT}_{\text{pure\_tr}}$, we have a \textit{weak tool} $t^{'}$ and a corresponding query $q'$. Using the synthesized data, we create:
\begin{compactenum} 
    \item $d_{\text{train}}^1 = ( T_s^{1'}, q_1, t_{1}^{'}, y_1^{1'}  )$, where the model needs to select \( t_{1}^{'} \), as only this weak tool is available.
    \item $d_{\text{train}}^2  = ( T_s^{1+}, q_1, t_1, y_1^1)$, where $T_s^{1+} = \{ t_1, t_{1}^{'}, \dots \}$, the model needs to select the strong tool $t_1$, with both weak and strong tools present.
\end{compactenum} 
By including $d_{\text{train}}^1$ and $d_{\text{train}}^2$ in the training set, the model learns to prioritize strong tools over weak ones, simulating \textit{weak-to-strong} generalization.



\subsection{Test Instances Construction}
\label{sec:scenario_generation}
Building upon the four scenarios defined in Section \ref{sec2.2}, we now describe how to construct instances from $\text{QT}_{\text{test}}$ for each scenario. Note that the \texttt{generate\_response} tool remains available across all scenarios.




\paragraph{\textit{Seen\_Query\_Unseen\_Tool}}
\label{para:cluster_2}
For a given query-tool cluster $C_2 = \{ q_2, q_2^{'}, t_2, t_{2'} \} \in \text{QT}_{\text{test}}$, we first construct a training instance with no available tools:
\[
d_{\text{train}}^1: (T_s^{N}, q_2, \text{None}, y_N)
\]
where the model must use \texttt{generate\_response}. For testing, we introduce unseen tools and construct:
\begin{align*}
d_{test}^{1}: (T_s^{2}, q_2, t_2, y_2^2), \\
d_{test}^{2}: (T_s^{2'}, q_2, t_{2}^{'}, y_2^{2'})
\end{align*}

\label{para:cluster_3}
Additionally, we use another cluster $C_3 = (q_3, q_3^{'}, t_3, t_{3}^{'}) \in \text{QT}_{\text{test}}$ to create a training instance with only weak tools:
\[
d_{train}^{1}: (T_s^{3'}, q_3, t_{3}^{'}, y_3^{3'})
\]
and test with both weak and strong tools:
\[
d_{test}^{1}: (T_s^{3+}, q_3, t_3, y_3^3)
\]
In this case, the strong tool remains unseen in the training data, 


\paragraph{Unseen\_Query\_Unseen\_Tool}
Using cluster $C_3$, we extend testing to unseen query and unseen tool:
\[
d_{test}^{1}: (T_s^{3}, q_3^{'}, t_3, y_{3'}^{3})
\]
where both $q_3^{'}$ and $t_3$ are absent from training data.




\paragraph{\textit{Seen\_Query\_Seen\_Tool}}
\label{para:cluster_4}
Using cluster $C_4 = \{q_4, q_4^{'}, t_4, t_4^{'}\} \in \text{QT}_{\text{test}}$, we construct training instances:
\begin{align*}
d_{train}^{1}: (T_s^{4'}, q_4, t_{4}^{'}, y_4^{4'}), \\
d_{train}^{2}: (T_s^{4}, q_{4'}, t_4, y_{4'}^{4})
\end{align*}
and test with expanded toolsets:
\[
d_{test}^{1}: (T_s^{4+}, q_{4}, t_4, y_{4}^{4})
\]



\paragraph{\textit{Unseen\_Query\_Seen\_Tool}}
Using cluster $C_3$, we construct test instances:
\begin{align*}
d_{test}^{1}: (T_s^{3'}, q_{3'}, t_{3}^{'}, y_{3'}^{3'}) \\
d_{test}^{2}: (T_s^{3'}, q_{3'}, \text{None}, y_N)
\end{align*}
where $q_{3'}$ is unseen while the tools have been exposed during training.

\subsection{Two-stage Output Training}
Efficient tool usage in real-world applications requires prioritizing tools in the toolset. Existing fine-tuning approaches often directly output the selected tool without considering their relative priority. For instance, even when the gold tool is included in the toolsets, \texttt{generate\_response} still take precedence over other irrelevant tools. To address this, we propose a two-stage process:
(1) Generating a ranked list of tools in the toolset.
(2) Providing detailed calling information for the highest-ranked tool.

Our ranking system considers three distinct cases:
\begin{compactenum}
    \item \textbf{No useful tools:} \texttt{generate\_response} ranks first, followed by other tools.
    \item \textbf{Single useful tool:} useful tool $>$ \texttt{generate\_response} $>$ others.
    \item \textbf{Multiple tools:} strong tool $>$ weak tool $>$ \texttt{generate\_response} $>$ other tools.
\end{compactenum}
The detailed prompt design for this task is available in Appendix~\ref{our_ex_app}.

\section{Experimental Setup and Dataset}
\subsection{Evaluation Metrics}
We evaluate model performance through a comprehensive assessment of tool invocation accuracy, comparing predicted outputs against ground truth across four critical dimensions: tool selection accuracy, parameter name identification, parameter value matching, and syntactic format correctness. For detailed explanations, please refer to App. \ref{app_eval}.

\subsection{Baseline}

We evaluate our approach using four models: \textbf{Llama-3.1-8B-Instruct}, \textbf{Llama-3.2-1B-Instruct} \cite{dubey2024llama3herdmodels}, \textbf{Mistral-Instruct-7B-v0.3} \cite{jiang2023mistral7b}, and \textbf{Phi-3.5-3B} \cite{abdin2024phi3technicalreporthighly}. For simplicity, we omitted the name ``instruct'' in some experimental results. We compare two types of methods: \textbf{tuning-free} approaches using GPT-3.5, GPT-4o-mini, GPT-4o \cite{openai2024gpt4technicalreport}, ToolLlama \cite{qin2024toolllm}, and GPT4Tools \cite{yang2023gpttools}; and \textbf{tuning-based} methods including \textbf{Original:fine-tuning on seed data} (cf. Section \ref{section3.1}) and \textbf{Half-Sample: using single examples from paired data} (e.g., selecting either the ``zero'' or ``one'' example from zero-to-one pairs). For model implementation details and data statistics, please see App.\ref{our_ex_app}.




\section{Experimental Results}
\begin{table}[!th]
\centering
\tiny
\begin{tabular}{l|c|c|c|c}
\hline
\multirow{2}{*}{\textbf{Method}} & \multirow{2}{*}{\textbf{Tool Selection}} & \multicolumn{2}{c|}{\textbf{Parameter}} & \multirow{2}{*}{\textbf{Format Accuracy}} \\ \cline{3-4}&& \textbf{Name}& \textbf{Value}       &\\ \hline

\rowcolor[gray]{0.9} 
\multicolumn{5}{c}{\textbf{Baselines}} \\ \hline
GPT-3.5       & 48.61 & 70.34 & 59.25 & 99.30 \\ \hline
GPT-4o Mini    & 58.80 & 78.86 & 69.05 & 99.93 \\ \hline
GPT-4o         & 75.87 & 72.00 & 64.09 & 99.81 \\ \hline
GPT4Tools              & 22.34 & 28.19 & 22.47 & 99.53 \\ \hline
ToolLLaMA                 & 17.73 & 24.59 & 18.43 & 99.38 \\ \hline
\rowcolor[gray]{0.9} 
\multicolumn{5}{c}{\textbf{Llama-3.2-1B-Instruct }} \\ \hline
Zero-shot        & 29.02 &  0.08 &  0.07 & 40.12 \\ \hline
Original & 31.63 & 35.67 & 31.12 & 99.88 \\ \hline
Half-Sample  & 61.52 & 58.48 & 52.19 & 99.92 \\ \hline
\textbf{GenTool }   & 86.31 & 76.92 & 68.81 & 99.88 \\ \hline
\rowcolor[gray]{0.9} 
\multicolumn{5}{c}{\textbf{Phi-3.5-3B-Instruct }} \\ \hline
Zero-shot          & 24.15 & 30.06 & 19.14 & 68.19 \\ \hline
Original   & 55.49 & 58.58 & 52.63 & 99.90 \\ \hline
Half-Sample    & 80.77 & 75.23 & 68.22 & 99.85 \\ \hline
\textbf{GenTool }    & 87.37 & 83.48 & 75.62 & 99.83 \\ \hline
\rowcolor[gray]{0.9} 
\multicolumn{5}{c}{\textbf{Mistral-Instruct-7B-v0.3 }} \\ \hline
Zero-shot      & 45.33 & 66.27 & 56.07 & 97.89 \\ \hline
Original & 18.28 & 19.71 & 17.18 & 99.25 \\ \hline
Half-Sample   & 51.00 & 49.74 & 44.52 & 99.78 \\ \hline
\textbf{GenTool }   & 80.28 & 80.30 & 72.43 & 99.68 \\ \hline
\rowcolor[gray]{0.9} 
\multicolumn{5}{c}{\textbf{LLaMA-3.1-8B-Instruct }} \\ \hline
Zero-shot        & 40.52 & 60.71 & 50.27 & \textbf{99.96} \\ \hline
Original& 48.09 & 42.67 & 38.03 & 99.82 \\ \hline
Half-Sample   & 60.47 & 73.39 & 66.53 & 99.72 \\ \hline
\textbf{GenTool }    & \textbf{90.15} & \textbf{86.05} & \textbf{77.64} & 99.45 \\ \hline
\end{tabular}
\caption{Overall tool-calling results. All baseline models are evaluated in a zero-shot setting. For different LLMs, the \textbf{original} and \textbf{half-sample} settings represent results obtained through fine-tuning.}
\label{tab:main_res}
\vspace{-0.5cm}
\end{table}


\subsection{Overall Performance}
Comprehensive evaluation results (Table \ref{tab:main_res})
show GenTool 's superior performance across all metrics. Models fine-tuned with GenTool consistently outperform GPT-based baselines, with particularly notable improvements in \textit{Tool Name Selection}. The LLaMA3.1-8B model fine-tuned with GenTool outperforms GPT-4o by 14.28\%, demonstrating GenTool 's strong potential in selecting the appropriate tool. This substantial gain suggests that our structured training approach better captures the underlying patterns of tool selection.

For \textit{tuning-free} methods, distinct performance patterns across model scales. Smaller models (LLama-3.1-1B, Phi-3.5-3B) struggle with instruction following, while models like GPT4Tools and ToolLLaMA achieve high format accuracy but lower performance than zero-shot baselines, probably due to domain shifts between their training datasets and UltraTool specific requirements, highlighting the challenges in cross-domain tool generalization.

For \textit{tuning-based} approaches, \textit{Half-Sample} outperforms \textit{Original}, highlighting the benefits of our training data. In addition, performance shows no consistent correlation with model size, indicating it might not be critical for tool generalization.



\begin{figure*}[!tpb]
    \centering
    \includegraphics[width=1\linewidth]{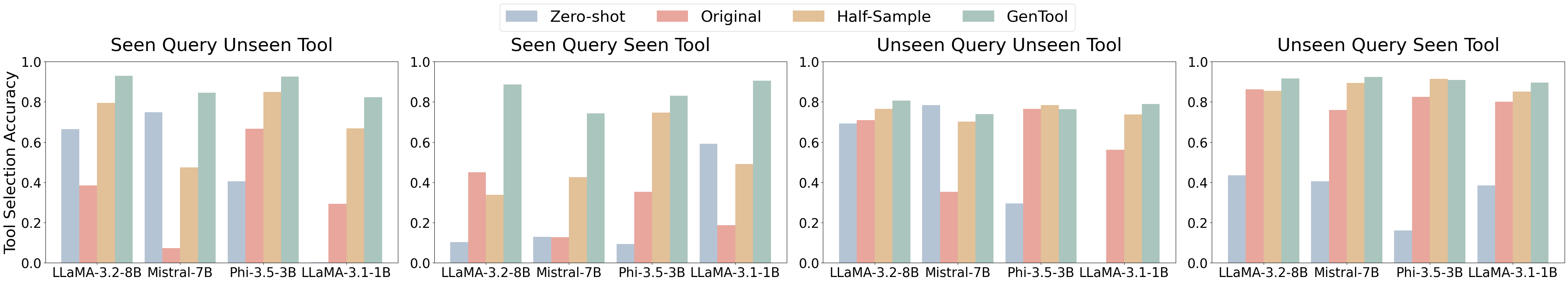}
    \caption{Detailed results for different test set scenarios. GenTool consistently outperforms all baselines. 
    }
    \label{scen4}
    
    \vspace{-0.4cm}
\end{figure*}

\subsection{Performance across Different Generalization Scenarios}
We examine GenTool 's performance across different generalization scenarios, excluding 
\textit{tuning-free baselines} as they treat all inputs as unseen. Figure \ref{scen4} shows that GenTool excels particularly with seen queries. For the \textit{Tool Selection} task, our GenTool demonstrates substantial improvements over the strongest baseline, achieving average gains of 18.4\% in the \textit{Seen\_Query\_Unseen\_Tool} and 34.1\% \textit{Seen\_Query\_Seen\_Tool} scenarios. 

The framework maintains strong performance even in challenging unseen query scenarios.
In the \textit{Unseen\_Query\_Unseen\_Tool} and \textit{Unseen\_Query\_Seen\_Tool} scenarios, GenTool  consistently outperforms all baselines. Our approach outperforms the best baseline (half-sample) by 2.8\% and 3.2\% respectively.  This performance suggests successful mitigation of overfitting to seen queries, a common challenge in tool learning systems.


\begin{table}[!t]
\scriptsize
\centering
\renewcommand{\arraystretch}{1.2}
\setlength{\tabcolsep}{6pt}
\begin{tabular}{l|c|c|c}
\hline
\multirow{2}{*}{\textbf{Model}}                 & \multirow{2}{*}{\textbf{Tool Selection}} & \multicolumn{2}{c}{\textbf{Parameter}} \\ \cline{3-4}
                                &                         & \textbf{Name}        & \textbf{Value}        \\ \hline
LLama-3.1-8B (Ours)                & \textbf{90.15}                   & \textbf{86.05}               & \textbf{77.64}                \\ 
-- w/o Rank                    & 83.68                   & 84.58               & 76.32                \\ \hline
Mistral-7B (Ours)              & \textbf{80.28}                   & \textbf{80.30}               & \textbf{72.43}                \\ 
-- w/o Rank                    & 69.39                   & 71.49               & 63.98                \\ \hline
\end{tabular}
\caption{Ablation Study Results: Impact of Removing the Tool Ranking Task During Fine-Tuning. We instruct the model to directly output the most suitable tool-calling information.}
\label{tab:ablation_study_inline_no_parse}
\vspace{-0.3cm}
\end{table}
\begin{figure}[!tpb]
    \centering
    \includegraphics[width=0.8\linewidth]{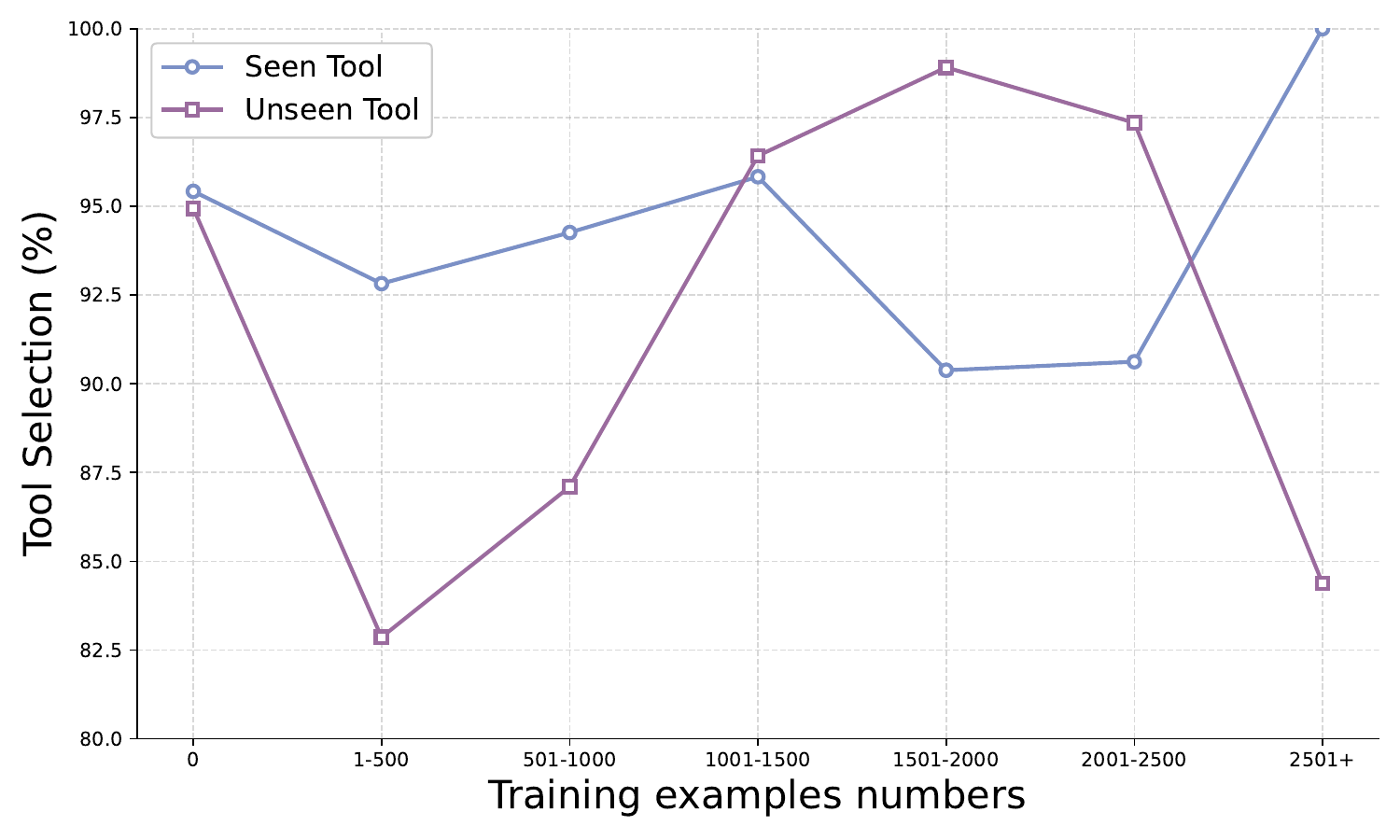}
    \caption{
    The relationship between the number of training examples related to the test set's gold tool and the test set's accuracy. The \textcolor{blue}{
    blue} line represents seen tools, while the \textcolor{violet}{purple} line denotes unseen tools.
    }
   
    \label{embedding}
    \vspace{-0.6cm}
\end{figure}
\subsection{Ablation Study}
To validate the effectiveness of our ranking mechanism, we conduct systematic ablation experiments examining its contribution to overall performance. Table \ref{tab:ablation_study_inline_no_parse} reveals that removing the ranking component leads to consistent performance degradation across all metrics. The impact is particularly obvious in Tool Selection accuracy, where LLama-3.1-8B and Mistral-7B models show significant decreases of 6.47\% and 10.89\%, respectively.


\section{Empirical Analysis}
In this section, we conduct comprehensive experiments to understand the factors influencing tool generalization and the limitations of models. Our analysis focuses on two key aspects: examining how different training data compositions affect model performance and investigating the fundamental characteristics of model capabilities. 

\subsection{Impact of Related Examples Quantity}
To investigate how related training examples influence model performance, we analyze the correlation between the number of related examples and test performance. We hypothesize that performance improves with increased exposure to related tools in the training set. Using the text-embedding-3-large model \cite{OpenAI2023EmbeddingModel}, we calculate semantic similarities for embeddings between gold tools in the training and test sets. For a test tool $i$, if a training tool $j$ has a cosine similarity greater than 0.5, $j$ is considered related to $i$.



Figure \ref{embedding} reveals distinct trends for Seen and Unseen Tools. For Unseen Tools, we observe a non-monotonic trend: performance initially declines but improves as related examples increase, suggesting an interaction between memorization and generalization.
In contrast, Seen Tools demonstrate consistent improvement with additional related examples, indicating effective knowledge transfer within similar tool categories.


\subsection{Contribution of Different Pair Types}
\begin{table}[t!]
\tiny
\centering
\begin{tabular}{l|c|c|c}
\hline
\multirow{2}{*}{\textbf{Model}}                 & \multirow{2}{*}{\textbf{Tool Selection}} & \multicolumn{2}{c}{\textbf{Parameter}} \\ \cline{3-4}
                                &                         & \textbf{Name}        & \textbf{Value}        \\ \hline
GenTool  & \textbf{90.15} & \textbf{86.05} & \textbf{77.64} \\ \hline
Weak-to-Strong & 63.36 & 82.72 & 73.65  \\ \hline
Zero-to-One & 71.01 & 62.37 & 55.87  \\ \hline
\end{tabular}
\caption{Results of the LLaMA-3.1-8B-Instruct model trained in a single generalization simulation scenario. \textbf{Weak-to-strong} indicates the model is trained only on the weak-to-strong-related dataset, while \textbf{zero-to-one} indicates the model is trained only on the zero-to-one-related dataset.}
\label{tab:transfer}
\vspace{-0.3cm}
\end{table}

We evaluate the individual impact of \textit{Zero-to-One} and \textit{Weak-to-Strong} pairs on model generalization by fine-tuning models exclusively on either Zero-to-One or Weak-to-Strong pairs. Table \ref{tab:transfer} shows that \textit{weak-to-strong} pairs enhance parameter name accuracy but lead to lower tool selection accuracy, excelling when tools are available but struggling to identify when no tool is applicable. In contrast, fine-tuned models with \textit{zero-one-one} pairs exhibit higher tool selection accuracy, indicating better capability in identifying scenarios where tool usage is inappropriate. This suggests both pair types are essential for comprehensive tool generalization.

\subsection{Training with Test-Related Examples Only}
To examine whether models understand tools during learning, we train models exclusively on test-relevant tools and queries.
Table \ref{tab:train_test}
reveals significant performance deterioration across all models, especially fine-tuned models. For instance, Mistral-7B's accuracy in \textit{Tool Selection} drops from 45.33\% to 11.56\%. These results indicate that models may not fully understand tools or queries, and slight variations in tool usage scenarios lead to sharp accuracy declines. This aligns with observations in RoTtuning \cite{ye-etal-2024-rotbench}.

\begin{table}[t!]
\scriptsize
\centering
\definecolor{specialblue}{RGB}{29, 66, 168}
\definecolor{specialgreen}{RGB}{27, 143, 34}

\begin{tabular}{l|c|c|c}
\hline
\multirow{2}{*}{\textbf{Model}} & \multirow{2}{*}{\textbf{Tool Selection}} & \multicolumn{2}{c}{\textbf{Parameter}}  \\ \cline{3-4} &  & \textbf{Name}& \textbf{Value}   \\ \hline
Llama-3.2 & 16.32 {\color{specialblue}(-69.99)} & 18.55 {\color{specialblue}(-58.37)} & 16.43 {\color{specialblue}(-52.38)}  \\ \hline
Phi-3.5-3B & \textbf{27.99} {\color{specialblue}(-59.38)} & \textbf{35.91} {\color{specialblue}(-47.57)} & \textbf{32.07} {\color{specialblue}(-43.55)} \\ \hline
Mistral-7B & 11.56 {\color{specialblue}(-68.72)} & 16.36 {\color{specialblue}(-63.94)} & 14.32 {\color{specialblue}(-58.11)} \\ \hline
Llama-3.1-8B & 17.11 {\color{specialblue}(-73.04)} & 24.57 {\color{specialblue}(-61.48)} & 21.51 {\color{specialblue}(-56.13)} \\ \hline
\end{tabular}
\caption{Performance comparison of models trained exclusively on test-relevant examples. {\color{specialgreen}{green}} and {\color{specialblue}{blue}} values represent the differences compared to the GenTool method respectively.}
\label{tab:train_test}
\end{table}
\subsection{Top Ranking Performance}
\begin{table}[!t]
\footnotesize
\centering
\begin{tabular}{l|c|c}
\hline
\textbf{Model} & (1) & (2) \\ \hline
Llama-3.2-1B-Instruct & 99.81 & 78.45 \\ \hline

Phi-3.5-3B & \textbf{99.82} & 78.56 \\ \hline
Mistral-Instruct-7B-v0.3 & 99.53 & 71.48 \\ \hline

Llama-3.1-8B-Instruct & 99.41 & \textbf{80.09} \\ \hline
\end{tabular}
\caption{Tool ranking analysis: (1) consistency between top-ranked in the first task and re-predicted tools in the second task; (2) accuracy of useful\/irrelevant tool ranking relative to \texttt{generate\_response}.}
\label{tab:rank_analysis}
\vspace{-0.4cm}
\end{table}

Our ranking-based mechanism requires models to rank tools and generate detailed calls for the top-ranked option. As shown in Table \ref{tab:rank_analysis}, fine-tuned models achieve nearly 100\% accuracy in executing
the top-ranked tool, but struggle to distinguish useful tools from irrelevant ones, with consistency below 80\%. This highlights a challenge in filtering irrelevant tools despite strong \textit{Tool Selection} accuracy.




\label{analysis_train_only}


\section{Related Work}
\subsection{Tool Learning with LLMs}  
Recent studies have enhanced large language models (LLMs) with tools such as code-related APIs \cite{patil2023gorilla}, mathematical functions \cite{gou2024critic}, and feedback mechanisms to refine outputs and reduce hallucinations \cite{dhuliawala-etal-2024-chain}. For example, \citet{mekala-etal-2024-toolverifier} emphasizes tool-based verification for reliable answers, while \citet{wang-etal-2024-llms-imaginarium} uses simulated errors to guide tool learning. \citet{Gao_Shi_Zhu_Fang_Xin_Ren_Chen_Ma_Ren_2024} propose progressive learning to improve tool usage across tasks. However, these approaches lack a comprehensive evaluation of tool generalization. Our work addresses this by defining the tool generalization problem and incorporating diverse scenarios into training.


\subsection{Synthetic Data Generation}
LLMs' generative capabilities have facilitated synthetic data generation, reducing annotation costs while maintaining consistency \cite{zelikman2022star, honovich-etal-2023-unnatural}. For tool learning, \citet{huang-etal-2024-planning-creation} and \citet{huang2024metatool} generate complex queries to evaluate tool usage, while \citet{qin2024toolllm} and \citet{tang2023toolalpaca} use APIs and instruction-response pairs to improve fine-tuning. However, these methods often overlook scenario complexity, limiting generalization performance. Our approach extends these methods by synthesizing data tailored to enhance performance across diverse tool generalization scenarios.




\section{Conclusion}
This work addresses the generalization challenges of LLMs in tool learning by proposing GenTool, a novel approach simulating the generalization process from two dimensions: \textit{Zero-to-One} and \textit{Weak-to-Strong Generalization}. GenTool further enhances LLMs’ tool selection by guiding them to output ranked tool lists.

Extensive experiments demonstrate that our method outperforms GPT-4o, achieving superior results across all generalization scenarios when compared to strong tuning-based and tuning-free baselines. Further analysis not only highlight that GenTool consistently delivers better performance but also reveal a key insight: existing LLMs rely heavily on memorizing tools rather than understanding their use. 
This indicates a significant gap between current LLMs and the goal of developing truly intelligent agents capable of effective tool utilization. 
We hope this research inspires further advancements in LLMs' tool-learning capabilities.



\section*{Limitation}
\paragraph{Model Scale}
Our experimental scope was constrained to backbone models under 8 billion parameters due to computational limitations. While GenTool has demonstrated promising results with these smaller models, its effectiveness with larger-scale architectures remains unexplored. This limitation is particularly noteworthy given that our training data was synthesized using GPT-4o. An important avenue for future research would be investigating whether GenTool's approach can enhance the fine-tuning capabilities of larger models, including GPT-4o itself.

\paragraph{Focusing on Single-query and Single-tool Scenarios}
Our work currently addresses single-query, single-tool scenarios, which represent 44.26\% of real-world applications according to the ToolBench dataset \cite{qin2024toolllm}. While this focus enabled us to conduct \textbf{the first} comprehensive investigation of tool learning generalization across two dimensions examining four distinct scenarios: \textit{seen\_query\_unseen\_tool}, \textit{seen\_query\_seen\_tool}, \textit{unseen\_query\_unseen\_tool}, and \textit{unseen\_query\_seen\_tool}, we acknowledge that real-world applications often require more complex multi-step planning and tool combinations. Our work establishes a foundation for understanding generalization in simpler contexts, paving the way for future research into more complex orchestration challenges involving multiple queries and tools.


\paragraph{Bias of Synthetic Data}
While using a single LLM for dataset construction is common in related research \cite{huang2024metatool,Gao_Shi_Zhu_Fang_Xin_Ren_Chen_Ma_Ren_2024}, our choice of GPT-4o for data synthesis may introduce inherent biases. However, although utilizing multiple LLMs could potentially reduce these biases, it would likely introduce variance in generation quality, potentially compromising evaluation stability. Given that our task primarily involves generating straightforward tools and queries, and that GPT-4o's biases are not directly related to the tool learning generalization problem, we determined that using a single, high-quality model for data synthesis was the most appropriate approach for this work.

\bibliography{acl_latex}

\begin{thebibliography}{29}
\providecommand{\natexlab}[1]{#1}

\bibitem[{Abdin et~al.(2024)Abdin, Aneja, Awadalla, Awadallah, Awan, Bach, Bahree, Bakhtiari, Bao, Behl, Benhaim, and Misha Bilenko~et}]{abdin2024phi3technicalreporthighly}
Marah Abdin, Jyoti Aneja, Hany Awadalla, Ahmed Awadallah, Ammar~Ahmad Awan, Nguyen Bach, Amit Bahree, Arash Bakhtiari, Jianmin Bao, Harkirat Behl, Alon Benhaim, and .~al Misha Bilenko~et. 2024.
\newblock \href {https://arxiv.org/abs/2404.14219} {Phi-3 technical report: A highly capable language model locally on your phone}.
\newblock \emph{Preprint}, arXiv:2404.14219.

\bibitem[{Deng et~al.(2023)Deng, Gu, Zheng, Chen, Stevens, Wang, Sun, and Su}]{deng2023mind2web}
Xiang Deng, Yu~Gu, Boyuan Zheng, Shijie Chen, Samuel Stevens, Boshi Wang, Huan Sun, and Yu~Su. 2023.
\newblock \href {https://arxiv.org/abs/2306.06070} {Mind2web: Towards a generalist agent for the web}.
\newblock \emph{Preprint}, arXiv:2306.06070.

\bibitem[{Dhuliawala et~al.(2024)Dhuliawala, Komeili, Xu, Raileanu, Li, Celikyilmaz, and Weston}]{dhuliawala-etal-2024-chain}
Shehzaad Dhuliawala, Mojtaba Komeili, Jing Xu, Roberta Raileanu, Xian Li, Asli Celikyilmaz, and Jason Weston. 2024.
\newblock \href {https://doi.org/10.18653/v1/2024.findings-acl.212} {Chain-of-verification reduces hallucination in large language models}.
\newblock In \emph{Findings of the Association for Computational Linguistics: ACL 2024}, pages 3563--3578, Bangkok, Thailand. Association for Computational Linguistics.

\bibitem[{Dubey et~al.(2024)Dubey, Jauhri, Pandey, Kadian, Al-Dahle, Letman, Mathur, Schelten, Yang, Fan, Goyal, Hartshorn, Yang, Mitra, Sravankumar, Korenev, Hinsvark, Rao, and et~al.}]{dubey2024llama3herdmodels}
Abhimanyu Dubey, Abhinav Jauhri, Abhinav Pandey, Abhishek Kadian, Ahmad Al-Dahle, Aiesha Letman, Akhil Mathur, Alan Schelten, Amy Yang, Angela Fan, Anirudh Goyal, Anthony Hartshorn, Aobo Yang, Archi Mitra, Archie Sravankumar, Artem Korenev, Arthur Hinsvark, Arun Rao, and et~al. 2024.
\newblock \href {https://arxiv.org/abs/2407.21783} {The llama 3 herd of models}.
\newblock \emph{Preprint}, arXiv:2407.21783.

\bibitem[{Gao et~al.(2024)Gao, Shi, Zhu, Fang, Xin, Ren, Chen, Ma, and Ren}]{Gao_Shi_Zhu_Fang_Xin_Ren_Chen_Ma_Ren_2024}
Shen Gao, Zhengliang Shi, Minghang Zhu, Bowen Fang, Xin Xin, Pengjie Ren, Zhumin Chen, Jun Ma, and Zhaochun Ren. 2024.
\newblock \href {https://doi.org/10.1609/aaai.v38i16.29759} {Confucius: Iterative tool learning from introspection feedback by easy-to-difficult curriculum}.
\newblock \emph{Proceedings of the AAAI Conference on Artificial Intelligence}, 38(16):18030--18038.

\bibitem[{Gou et~al.(2024)Gou, Shao, Gong, yelong shen, Yang, Duan, and Chen}]{gou2024critic}
Zhibin Gou, Zhihong Shao, Yeyun Gong, yelong shen, Yujiu Yang, Nan Duan, and Weizhu Chen. 2024.
\newblock \href {https://openreview.net/forum?id=Sx038qxjek} {{CRITIC}: Large language models can self-correct with tool-interactive critiquing}.
\newblock In \emph{The Twelfth International Conference on Learning Representations}.

\bibitem[{Honovich et~al.(2023)Honovich, Scialom, Levy, and Schick}]{honovich-etal-2023-unnatural}
Or~Honovich, Thomas Scialom, Omer Levy, and Timo Schick. 2023.
\newblock \href {https://doi.org/10.18653/v1/2023.acl-long.806} {Unnatural instructions: Tuning language models with (almost) no human labor}.
\newblock In \emph{Proceedings of the 61st Annual Meeting of the Association for Computational Linguistics (Volume 1: Long Papers)}, pages 14409--14428, Toronto, Canada. Association for Computational Linguistics.

\bibitem[{Hsieh et~al.(2023)Hsieh, Chen, Li, Fujii, Ratner, Lee, Krishna, and Pfister}]{hsieh2023tooldocumentationenableszeroshot}
Cheng-Yu Hsieh, Si-An Chen, Chun-Liang Li, Yasuhisa Fujii, Alexander Ratner, Chen-Yu Lee, Ranjay Krishna, and Tomas Pfister. 2023.
\newblock \href {https://arxiv.org/abs/2308.00675} {Tool documentation enables zero-shot tool-usage with large language models}.
\newblock \emph{Preprint}, arXiv:2308.00675.

\bibitem[{Huang et~al.(2024{\natexlab{a}})Huang, Zhong, Lu, Zhu, Gao, Liu, Hou, Zeng, Wang, Shang, Jiang, Xu, and Liu}]{huang-etal-2024-planning-creation}
Shijue Huang, Wanjun Zhong, Jianqiao Lu, Qi~Zhu, Jiahui Gao, Weiwen Liu, Yutai Hou, Xingshan Zeng, Yasheng Wang, Lifeng Shang, Xin Jiang, Ruifeng Xu, and Qun Liu. 2024{\natexlab{a}}.
\newblock \href {https://doi.org/10.18653/v1/2024.findings-acl.259} {Planning, creation, usage: Benchmarking {LLM}s for comprehensive tool utilization in real-world complex scenarios}.
\newblock In \emph{Findings of the Association for Computational Linguistics: ACL 2024}, pages 4363--4400, Bangkok, Thailand. Association for Computational Linguistics.

\bibitem[{Huang et~al.(2024{\natexlab{b}})Huang, Jung, Kumar, Kachuee, Li, Xu, and Chen}]{huang-etal-2024-planning}
Tenghao Huang, Dongwon Jung, Vaibhav Kumar, Mohammad Kachuee, Xiang Li, Puyang Xu, and Muhao Chen. 2024{\natexlab{b}}.
\newblock \href {https://doi.org/10.18653/v1/2024.findings-naacl.61} {Planning and editing what you retrieve for enhanced tool learning}.
\newblock In \emph{Findings of the Association for Computational Linguistics: NAACL 2024}, pages 975--988, Mexico City, Mexico. Association for Computational Linguistics.

\bibitem[{Huang et~al.(2024{\natexlab{c}})Huang, Shi, Li, Fan, Wu, Zhang, Liu, Zhou, Wan, Gong, and Sun}]{huang2024metatool}
Yue Huang, Jiawen Shi, Yuan Li, Chenrui Fan, Siyuan Wu, Qihui Zhang, Yixin Liu, Pan Zhou, Yao Wan, Neil~Zhenqiang Gong, and Lichao Sun. 2024{\natexlab{c}}.
\newblock \href {https://openreview.net/forum?id=R0c2qtalgG} {Metatool benchmark for large language models: Deciding whether to use tools and which to use}.
\newblock In \emph{The Twelfth International Conference on Learning Representations}.

\bibitem[{Iskander et~al.(2024)Iskander, Tolmach, Shapira, Cohen, and Karnin}]{iskander-etal-2024-quality}
Shadi Iskander, Sofia Tolmach, Ori Shapira, Nachshon Cohen, and Zohar Karnin. 2024.
\newblock \href {https://aclanthology.org/2024.emnlp-main.285} {Quality matters: Evaluating synthetic data for tool-using {LLM}s}.
\newblock In \emph{Proceedings of the 2024 Conference on Empirical Methods in Natural Language Processing}, pages 4958--4976, Miami, Florida, USA. Association for Computational Linguistics.

\bibitem[{Jiang et~al.(2023)Jiang, Sablayrolles, Mensch, Bamford, Chaplot, de~las Casas, Bressand, Lengyel, Lample, Saulnier, Lavaud, Lachaux, Stock, Scao, Lavril, Wang, Lacroix, and Sayed}]{jiang2023mistral7b}
Albert~Q. Jiang, Alexandre Sablayrolles, Arthur Mensch, Chris Bamford, Devendra~Singh Chaplot, Diego de~las Casas, Florian Bressand, Gianna Lengyel, Guillaume Lample, Lucile Saulnier, Lélio~Renard Lavaud, Marie-Anne Lachaux, Pierre Stock, Teven~Le Scao, Thibaut Lavril, Thomas Wang, Timothée Lacroix, and William~El Sayed. 2023.
\newblock \href {https://arxiv.org/abs/2310.06825} {Mistral 7b}.
\newblock \emph{Preprint}, arXiv:2310.06825.

\bibitem[{Kim et~al.(2023)Kim, Baldi, and McAleer}]{DBLP:journals/corr/abs-2303-17491}
Geunwoo Kim, Pierre Baldi, and Stephen McAleer. 2023.
\newblock \href {https://doi.org/10.48550/arXiv.2303.17491} {Language models can solve computer tasks}.
\newblock \emph{CoRR}, abs/2303.17491.

\bibitem[{Loshchilov and Hutter(2019)}]{loshchilov2018decoupled}
Ilya Loshchilov and Frank Hutter. 2019.
\newblock \href {https://openreview.net/forum?id=Bkg6RiCqY7} {Decoupled weight decay regularization}.
\newblock In \emph{International Conference on Learning Representations}.

\bibitem[{Mekala et~al.(2024)Mekala, Weston, Lanchantin, Raileanu, Lomeli, Shang, and Dwivedi-Yu}]{mekala-etal-2024-toolverifier}
Dheeraj Mekala, Jason~E Weston, Jack Lanchantin, Roberta Raileanu, Maria Lomeli, Jingbo Shang, and Jane Dwivedi-Yu. 2024.
\newblock \href {https://aclanthology.org/2024.findings-emnlp.289} {{TOOLVERIFIER}: Generalization to new tools via self-verification}.
\newblock In \emph{Findings of the Association for Computational Linguistics: EMNLP 2024}, pages 5026--5041, Miami, Florida, USA. Association for Computational Linguistics.

\bibitem[{OpenAI(2023)}]{OpenAI2023EmbeddingModel}
OpenAI. 2023.
\newblock New and improved embedding model.
\newblock \url{https://openai.com/blog/new-and-improved-embedding-model}.
\newblock Accessed: 2024-11-25.

\bibitem[{OpenAI et~al.(2024)OpenAI, Achiam, Adler, Agarwal, Ahmad, Akkaya, Aleman, Almeida, Altenschmidt, Altman, Anadkat, Avila, Babuschkin, and etc.}]{openai2024gpt4technicalreport}
OpenAI, Josh Achiam, Steven Adler, Sandhini Agarwal, Lama Ahmad, Ilge Akkaya, Florencia~Leoni Aleman, Diogo Almeida, Janko Altenschmidt, Sam Altman, Shyamal Anadkat, Red Avila, Igor Babuschkin, and Suchir~Balaji etc. 2024.
\newblock \href {https://arxiv.org/abs/2303.08774} {Gpt-4 technical report}.
\newblock \emph{Preprint}, arXiv:2303.08774.

\bibitem[{Paranjape et~al.(2023)Paranjape, Lundberg, Singh, Hajishirzi, Zettlemoyer, and Ribeiro}]{paranjape2023artautomaticmultistepreasoning}
Bhargavi Paranjape, Scott Lundberg, Sameer Singh, Hannaneh Hajishirzi, Luke Zettlemoyer, and Marco~Tulio Ribeiro. 2023.
\newblock \href {https://arxiv.org/abs/2303.09014} {Art: Automatic multi-step reasoning and tool-use for large language models}.
\newblock \emph{Preprint}, arXiv:2303.09014.

\bibitem[{Patil et~al.(2023)Patil, Zhang, Wang, and Gonzalez}]{patil2023gorilla}
Shishir~G. Patil, Tianjun Zhang, Xin Wang, and Joseph~E. Gonzalez. 2023.
\newblock Gorilla: Large language model connected with massive apis.
\newblock \emph{arXiv preprint arXiv:2305.15334}.

\bibitem[{Qin et~al.(2024)Qin, Liang, Ye, Zhu, Yan, Lu, Lin, Cong, Tang, Qian, Zhao, Hong, Tian, Xie, Zhou, Gerstein, dahai li, Liu, and Sun}]{qin2024toolllm}
Yujia Qin, Shihao Liang, Yining Ye, Kunlun Zhu, Lan Yan, Yaxi Lu, Yankai Lin, Xin Cong, Xiangru Tang, Bill Qian, Sihan Zhao, Lauren Hong, Runchu Tian, Ruobing Xie, Jie Zhou, Mark Gerstein, dahai li, Zhiyuan Liu, and Maosong Sun. 2024.
\newblock \href {https://openreview.net/forum?id=dHng2O0Jjr} {Tool{LLM}: Facilitating large language models to master 16000+ real-world {API}s}.
\newblock In \emph{The Twelfth International Conference on Learning Representations}.

\bibitem[{Qu et~al.(2024)Qu, Dai, Wei, Cai, Wang, Yin, Xu, and Wen}]{qu2024toollearninglargelanguage}
Changle Qu, Sunhao Dai, Xiaochi Wei, Hengyi Cai, Shuaiqiang Wang, Dawei Yin, Jun Xu, and Ji-Rong Wen. 2024.
\newblock \href {https://doi.org/10.1007/s11704-024-40678-2} {Tool learning with large language models: A survey}.
\newblock \emph{Preprint}, arXiv:2405.17935.

\bibitem[{Rasley et~al.(2020)Rasley, Rajbhandari, Ruwase, and He}]{10.1145/3394486.3406703}
Jeff Rasley, Samyam Rajbhandari, Olatunji Ruwase, and Yuxiong He. 2020.
\newblock \href {https://doi.org/10.1145/3394486.3406703} {Deepspeed: System optimizations enable training deep learning models with over 100 billion parameters}.
\newblock In \emph{Proceedings of the 26th ACM SIGKDD International Conference on Knowledge Discovery \& Data Mining}, KDD '20, page 3505–3506, New York, NY, USA. Association for Computing Machinery.

\bibitem[{Schick et~al.(2023)Schick, Dwivedi-Yu, Dessi, Raileanu, Lomeli, Hambro, Zettlemoyer, Cancedda, and Scialom}]{schick2023toolformer}
Timo Schick, Jane Dwivedi-Yu, Roberto Dessi, Roberta Raileanu, Maria Lomeli, Eric Hambro, Luke Zettlemoyer, Nicola Cancedda, and Thomas Scialom. 2023.
\newblock \href {https://openreview.net/forum?id=Yacmpz84TH} {Toolformer: Language models can teach themselves to use tools}.
\newblock In \emph{Thirty-seventh Conference on Neural Information Processing Systems}.

\bibitem[{Tang et~al.(2023)Tang, Deng, Lin, Han, Liang, and Sun}]{tang2023toolalpaca}
Qiaoyu Tang, Ziliang Deng, Hongyu Lin, Xianpei Han, Qiao Liang, and Le~Sun. 2023.
\newblock \href {https://arxiv.org/abs/2306.05301} {Toolalpaca: Generalized tool learning for language models with 3000 simulated cases}.
\newblock \emph{Preprint}, arXiv:2306.05301.

\bibitem[{Wang et~al.(2024)Wang, Fang, Eisner, Van~Durme, and Su}]{wang-etal-2024-llms-imaginarium}
Boshi Wang, Hao Fang, Jason Eisner, Benjamin Van~Durme, and Yu~Su. 2024.
\newblock \href {https://doi.org/10.18653/v1/2024.acl-long.570} {{LLM}s in the imaginarium: Tool learning through simulated trial and error}.
\newblock In \emph{Proceedings of the 62nd Annual Meeting of the Association for Computational Linguistics (Volume 1: Long Papers)}, pages 10583--10604, Bangkok, Thailand. Association for Computational Linguistics.

\bibitem[{Yang et~al.(2023)Yang, Song, Li, Zhao, Ge, Li, and Shan}]{yang2023gpttools}
Rui Yang, Lin Song, Yanwei Li, Sijie Zhao, Yixiao Ge, Xiu Li, and Ying Shan. 2023.
\newblock \href {https://openreview.net/forum?id=cwjh8lqmOL} {{GPT}4tools: Teaching large language model to use tools via self-instruction}.
\newblock In \emph{Thirty-seventh Conference on Neural Information Processing Systems}.

\bibitem[{Ye et~al.(2024)Ye, Wu, Gao, Huang, Li, Li, Fan, Zhang, Gui, and Huang}]{ye-etal-2024-rotbench}
Junjie Ye, Yilong Wu, Songyang Gao, Caishuang Huang, Sixian Li, Guanyu Li, Xiaoran Fan, Qi~Zhang, Tao Gui, and Xuanjing Huang. 2024.
\newblock \href {https://aclanthology.org/2024.emnlp-main.19} {{R}o{TB}ench: A multi-level benchmark for evaluating the robustness of large language models in tool learning}.
\newblock In \emph{Proceedings of the 2024 Conference on Empirical Methods in Natural Language Processing}, pages 313--333, Miami, Florida, USA. Association for Computational Linguistics.

\bibitem[{Zelikman et~al.(2022)Zelikman, Wu, Mu, and Goodman}]{zelikman2022star}
Eric Zelikman, Yuhuai Wu, Jesse Mu, and Noah Goodman. 2022.
\newblock \href {https://openreview.net/forum?id=_3ELRdg2sgI} {{ST}ar: Bootstrapping reasoning with reasoning}.
\newblock In \emph{Advances in Neural Information Processing Systems}.

\end{thebibliography}
\clearpage
\appendix
\begin{table*}[!t]
\footnotesize
\centering
\begin{tabular}{lcccc}

\toprule
\textbf{Method} & \textbf{Base Model} & \makecell{\textbf{Dataset} \\ \textbf{Construction}} & \makecell{\textbf{Generalization Scenairo} \\ \textbf{in Evaluation}} & \makecell{\textbf{Candidates} \\ \textbf{Construction}} \\

\midrule
Toolformer \cite{schick2023toolformer} & GPT-J-6B    & In-Context Learning        & 1   & Manually  \\
GPT4Tools \cite{yang2023gpttools}   & Vicuna-13B  & Manually                  & 1   & Manually  \\
ToolAlpaca \cite{tang2023toolalpaca}   & Vicuna-13B  & Simulation                  & 1 & Manually  \\
ToolBench \cite{qin2024toolllm}     & LLaMA-30B   & Self-Instruct                & 1& Manually  \\
ToolVerify \cite{mekala-etal-2024-toolverifier}& LLaMA-70B & Self-Instruct& 2  & Random \\
Confucius \cite{Gao_Shi_Zhu_Fang_Xin_Ren_Chen_Ma_Ren_2024}                            & LLaMA-7B    & Iterative Self-Instruct                     & 2  & Retrieval \\
STE \cite{wang-etal-2024-llms-imaginarium} & GPT-3.5 &Simulation& 1 &Manually   \\
\textbf{Ours} & GPT-4o & Simulation & 4 & Retrieval \\
\bottomrule
\end{tabular}
\caption{\textbf{Dataset construction} refers to the method used to obtain the training dataset.
\textbf{Generalization scenario in evaluation} indicates the number of generalization types covered during testing. \textbf{Candidates construction} specifies whether the candidate set is meticulously curated manually or generated using the same retrieval method as employed in real-world application scenarios.
}
\label{data_compare}
\end{table*}

\begin{table}[t!]
\footnotesize
    \centering
\begin{tabular}{c|ccc}
\hline
\textbf{Dataset} & \makecell{\textbf{Tools} \\ \textbf{amount}} & \makecell{\textbf{Instance} \\ \textbf{amount}} & \makecell{\textbf{Avg. word} \\ \textbf{input/output}} \\
\hline         Training &868&26,042&597/27 \\
        \hline
        Test cases (1) &471&2,996&610/33 \\
        Test cases (2)  &278&3,072&573/22 \\ 
        Test cases (3)  &252&448&574/32 \\
        Test cases (4)  &125&728&554/25  \\
        \hline
    \end{tabular}
    \caption{The data statistics for training and testing sets constructed from synthetic data, where the test set is divided into the following four categories based on generalization scenarios: (1) Seen\_query\_unseen\_tool (2) Seen\_query\_seen\_tool (3) Unseen\_query\_unseen\_tool
    (4) Unseen\_query\_seen\_tool}
    \label{tab:data_statistics}
\end{table}

\begin{table}[tbp]
\footnotesize
\centering
\begin{tabular}{p{0.25\textwidth}c}
\hline
\textbf{Quality Review Question} & \textbf{Correctness Rate} \\
\hline
Is the instruction a valid and well-formed query? & 94\% \\
\hline
Can the reference tool provide at least a partial solution to the query? & 97\% \\
\hline
Does the tool set selection comply with output specifications? (e.g., for ``none'' outputs, confirming no tool in the set can partially address the query) & 97\% \\
\hline
Are all tool call parameters correctly extracted or reasonably inferred from the instruction without missing or imagined values? & 96\% \\
\hline
All fields are valid & 93\% \\
\hline
\end{tabular}
\caption{Quality Review Results for synthetic data}
\label{tab:quality_review}
\end{table}

\begin{table*}[htbp]
\small
\centering
\begin{tabular}{p{0.95\textwidth}}
\hline

\multicolumn{1}{c}{\textbf{Tool Ranking and Invocation Process}} \\
\hline
You are a professional tool selection assistant. You will be given a query and a corresponding toolset. You have two consecutive tasks. \\
\textbf{First Task} \\
Rank the available tools in the toolset based on relevance. Output format is a list where each item is the name of the tool (from 'name' field in toolset list). Irrelevant tools should be placed after generate\_response(), relevant tools before generate\_response(). The order of irrelevant tools doesn't matter. Example format: [tool\_name\_1, generate\_response(), tool\_name\_2, tool\_name\_3, tool\_name\_4, tool\_name\_5] \\
\textbf{Second Task} \\
Invoke the tool ranked first from the previous task and fill required parameter values. Parameter format: "parameter\_name"="parameter\_value", with multiple parameters separated by commas. Parameter names from 'properties' field under 'arguments', parameter values from user's query. If generate\_response() ranks first, output generate\_response(). Example format: ["tool\_name\_1"("parameter\_name1"="parameter\_value1", "parameter\_name2"="parameter\_value2")] \\
\textbf{Final Output Format (JSON):} \\
\{ \\
\quad "The output of the first task": [], \\
\quad "The output of the second task": [] \\
\} \\
\textbf{Input:} \\
Query: \{input\_query\} \\
Toolset: \{tools\} \\
\hline
\end{tabular}
\caption{Prompt for the GenTool  methodology to rank and invoke the most relevant and effective tools from a given toolset}
\label{GenTool _prompt}
\end{table*}

\begin{table*}[htbp]
\small
\centering
\begin{tabular}{p{0.95\textwidth}}
\hline
\multicolumn{1}{c}{\textbf{Tool Generation Process}} \\
\hline
You are a professional tool creation assistant. Given a \#user query\#, a set of \#existing tools\#, your task is as follows: \\
1. Create a \#new tool\# to solve the \#user query\#. \\
2. The \#new tool\# you create should be less effective than the \#existing tools\# in answering the \#existing problem\#. \\
3. Effectiveness can be judged from the following perspectives: \\
\quad 3.1 The match between the tool names and the problem. \\
\quad 3.2 The length of the tool descriptions and their match with the problem. \\
\quad 3.3 The names, descriptions, and number of tool parameters and their match with the problem. \\
\quad 3.4 The tool's returned results and descriptions and their match with the problem. \\
4. The format of the \#new tool\# should be the same as the tools in the \#existing tools\#. \\
Below are examples of inputs for \#user query\#, and \#existing tools\#, and outputs for the \#new tool\#: \\
Please generate \#new tool\# in JSON format based on the following \#user query\# and \#existing tools\#. Ensure the JSON string format is correct and do not output anything else. \\
\textbf{\#user query\#:} \\
\{user\_query\} \\
\textbf{\#existing tools\#:} \\
\{ex\_tools\} \\
\textbf{\#new tool\#} \\
\hline
\end{tabular}
\caption{Prompt for using GPT-4o to generate new weak tool}
\label{tab:tool-generation}
\end{table*}

\begin{table*}[ht]
\small
\centering
\begin{tabular}{p{0.9\textwidth}}
\hline
\multicolumn{1}{c}{\textbf{Query Generation Process}} \\
\hline
You are a professional question generation assistant. Based on the \#weak tool sets\# and \#strong tool sets\#, your task is to generate 10 questions that are suitable for answering using the \#strong tool sets\#. The requirements are as follows: \\
The questions should be as detailed as possible. The questions should ideally only require the \#strong tool sets\# for answering, without needing assistance from other tools. Ensure that the examples provided are distinct from each other. You can use various sentence structures, such as commands or requests, and adjust the level of detail as needed. The questions can also be answered by the \#weak tool sets\#, but the \#weak tool sets\# should not match the questions as well as the \#strong tool sets\#. \\
Based on the following \#weak tool sets\# and \#strong tool sets\#, generate 10 different questions in list format, following the example format. Do not output anything else. \\
\textbf{\#weak tool sets\#:} \texttt{\{weak\}} \\
\textbf{\#strong tool sets\#:} \texttt{\{strong\}} \\
\textbf{\#output\#:} \\
\hline
\end{tabular}
\caption{Prompt designed for generating diverse new queries.}
\label{tab:query_generation_prompt}
\end{table*}

\begin{table*}[ht]
\small
\centering
\begin{tabular}{p{0.9\textwidth}}
\hline
\textbf{Calling Annotation Prompt:} \\
You are a professional tool matching assistant. Based on the \#user query\# and \#toolsets\#, your task is to perform tool matching for the \#user query\# and output a \#detailed execution plan\# with the following requirements: \\
In the \#detailed execution plan\#, provide the tool names and their specific parameters. The form of parameter passing should be: "parameter name: parameter value". Ensure that the parameter names used come from the defined parameter name set in the corresponding tools, and ensure that the parameter values have sources, which can come from two parts: user instructions or some user-related information (such as personal information like name, ID number, account, password, etc.). \\
Below are some examples of input \#user query\# and \#toolsets\# and the output \#detailed execution plan\#: \\
\textbf{\#user query\#:} \\
'Please create a new file named "2023 October Work Log.txt" with the initial entry "Tasks Completed Today"' \\
\textbf{\#toolsets\#:} \\
\{demons\_example\_tool\_set\} \\
\textbf{\#output\#:} \\
file\_write(file\_path='Desktop/2023 October Work Log.txt', content='Tasks Completed Today') \\
\\
Please generate a \#detailed execution plan\# in JSON format based on the following \#user query\# and \#toolsets\#, following the format of the examples. Ensure the JSON string format is correct and do not output anything else. \\
\textbf{\#user query\#:} \\
\texttt{\{query\}} \\
\textbf{\#toolsets\#:} \\
\texttt{\{tools\}} \\
\textbf{\#output\#:} \\
\hline
\end{tabular}
\caption{Prompt for generating a detailed calling information based on user query and toolsets.}
\label{tab:calling_annotation_prompt}
\end{table*}

\section{Dataset generation and checking}
\label{app_datasets}
\subsection{Dataset Generation Prompts}

\begin{figure*}[t]
    \centering
    \includegraphics[width=1\linewidth]{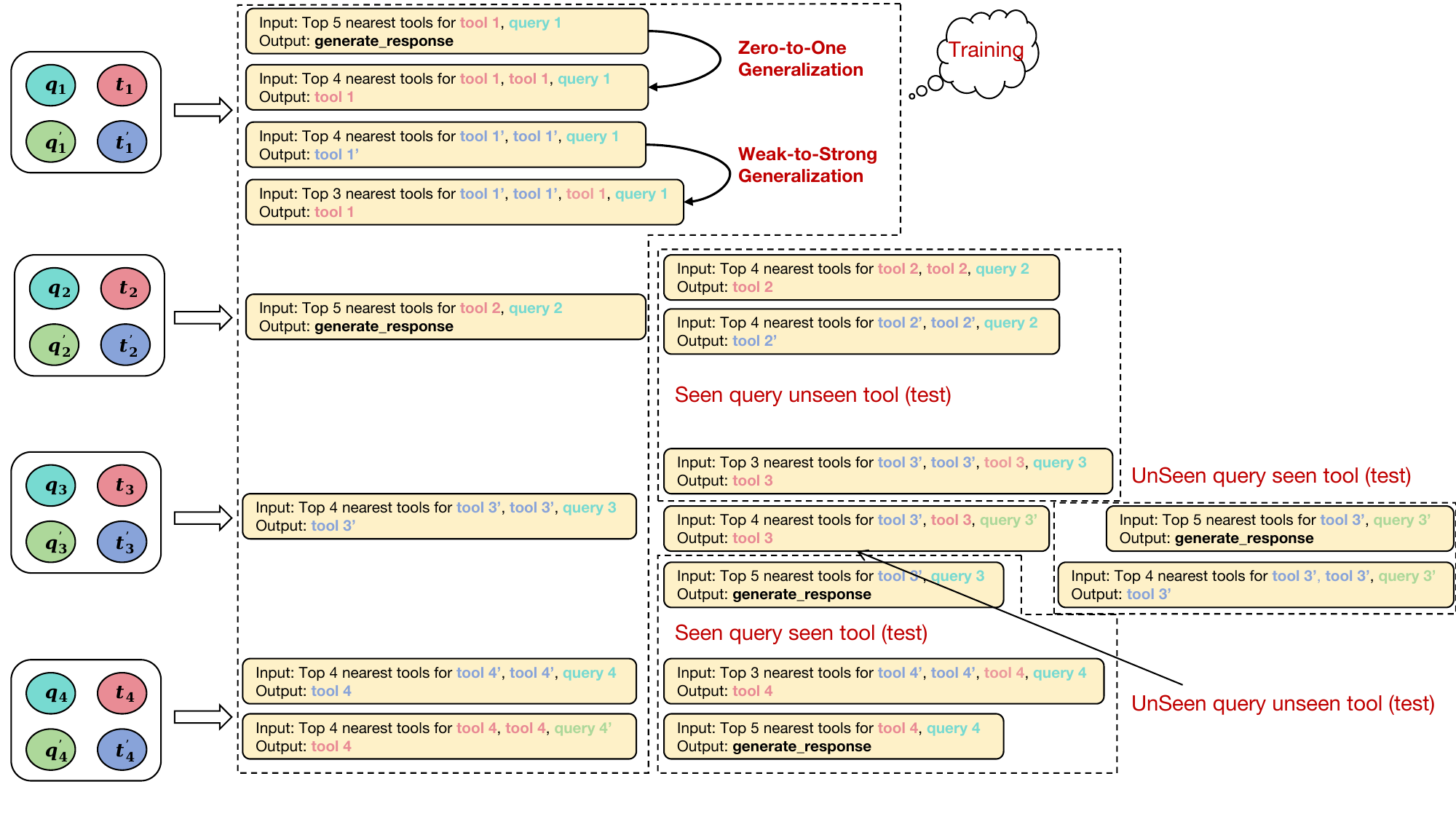}
    \caption{ 
    The detailed process of training and testing data division based on queries and the tool library: Different tool-query pairs play distinct roles. The frist cluster focuses on constructing the training set. The second cluster is dedicated to building the training and testing sets for \textit{seen\_query\_unseen\_tool}. The third cluster provides training and testing sets for \textit{seen\_query\_unseen\_tool}, \textit{unseen\_query\_unseen\_tool}, and \textit{unseen\_query\_seen\_tool}. Finally, the fourth cluster constructs both the training and testing sets for \textit{seen\_query\_seen\_tool}.
    }
    \label{fig:GenTool _eval_scenario}
\end{figure*}
\label{app_a}
Table~\ref{tab:tool-generation}, Table~\ref{tab:query_generation_prompt} and Table~\ref{tab:calling_annotation_prompt} present the prompts used to generate our dataset with GPT4o. Specifically:
\begin{itemize}
    \item Table~\ref{tab:tool-generation} contains prompts designed to generate \textit{weak tools} based on existing query-tool pairs.
    \item Table~\ref{tab:query_generation_prompt} includes prompts for diversifying query generation based on \textit{strong tools} and \textit{newly generated weak tools}.
    \item Table~\ref{tab:calling_annotation_prompt} provides prompts for generating output information based on query and tool combinations.
\end{itemize}

Figure ~\ref{fig:GenTool _eval_scenario} shows the example that we used in Section \ref{sec:scenario_generation} to explain  how we generate the four generalization scenarios.

\subsection{Human Verification}
\label{app_check}
To ensure data quality, we implement a systematic verification process. We randomly sample 200 instances, including queries, strong tools, the generated weak tools, and the queries derived from weak tools. An expert annotator, with extensive experience in API design and natural language processing, evaluates a random sample of 200 instances including queries, strong tools, the generated weak tools, and the queries derived from weak tools. Inspired by \citet{iskander-etal-2024-quality}, the verification process for each instance follow the criteria in Table \ref{tab:quality_review}.

Results in Table \ref{tab:quality_review} show that over 90\% of all fields are valid, significantly higher than the 33\% parameter alignment error rate observed in datasets like ToolBench and ToolAlpaca \cite{iskander-etal-2024-quality}. These high verification scores demonstrate the effectiveness of our generation pipeline in producing high-quality, reliable training data for tool generalization tasks.

\section{Dataset Comparison}


To further emphasize the distinctions between our work and existing approaches, Table~\ref{data_compare} presents a comparison with recent studies that leverage LLMs for synthetic data generation followed by fine-tuning. Notably, our method employs the state-of-the-art GPT-4o model for data generation, ensuring high-quality data. Moreover, we tackle diverse generalization scenarios, including complex compositional relationships between queries and tools, which remain unexplored in prior research. To further elevate the challenge of tool selection, we adopt a retrieval-based strategy to construct candidate toolsets, significantly testing the model’s ability to identify the correct tool. Our GenTool  method achieves an accuracy of up to 90.15\% on the tool selection task, demonstrating the effectiveness and superiority of our approach.

\section{Experiment Setup}
\label{our_ex_app}

\subsection{Data Statistics}
To validate the effectiveness of GenTool , we synthesized a dataset covering 22 domains, 1,633 tools, and 33,286 samples. We used 67\% of the tool clusters to construct training sets for \textit{zero-to-one} and \textit{weak-to-strong} generalization and 33\% of the clusters to simulate test scenarios. Detailed statistics are shown in Table~\ref{tab:data_statistics}.

\subsection{Implementation Details}
For our fine-tuning experiments, all model fine-tuning is done using the AdamW optimizer \cite{loshchilov2018decoupled} with \( \beta = (0.9, 0.999) \), \( \epsilon = 10^{-8} \), and no weight decay. The learning rate is set to 3e-5, and we use a linear scheduler with a warm-up ratio of 0.1. The batch size is set to 4, with a training epoch of 4, and training examples are truncated to 3076 tokens. We optimize the model training using the DeepSpeed zero strategy \cite{10.1145/3394486.3406703}. The model training can be completed in 15 hours using four Nvidia A100 PCIe 80GB GPUs. Notably, we exclude the result information from calling $T_{gold}$, as the primary focus of this study is on the tool generalization problem rather than the execution of actual API calls.  

In our fine-tuning process, we involve two tasks: \textit{Tool Ranking} and \textit{Tool Invocation}. The instructions are set as follows:
\begin{itemize}
    \item The input to the model follows the format: \texttt{[Input Query: ... | Toolset: ...]}.
    \item The model's gold output format is:
    \begin{quote}
        \texttt{The output of the first task is: [Tool1, Tool2, Tool3]. The output of the second task is: ...}.
    \end{quote}
\end{itemize}

Table~\ref{GenTool _prompt} shows the detailed prompt when we finetune the LLMs under the GenTool  training framework.


\begin{table*}[h!]
\centering
\begin{tabular}{l|c|c|c|c}
\hline
\multirow{2}{*}{\textbf{Method}} & \multirow{2}{*}{\textbf{Tool Selection}} & \multicolumn{2}{c|}{\textbf{Parameter}} & \multirow{2}{*}{\textbf{Format Accuracy}} \\ \cline{3-4}&& \textbf{Name}& \textbf{Value}       &\\ \hline
\multicolumn{5}{c}{\textbf{Seen Query, Unseen Tool}} \\ \hline
GenTool  & \textbf{92.84} & \textbf{85.27} & \textbf{76.55} & 98.96 \\ \hline
Weak-to-Strong & 90.21 & 83.80 & 75.82 & 99.57 \\ \hline
Zero-to-One & 63.32 & 57.18 & 50.81 & \textbf{99.77} \\ \hline
\multicolumn{5}{c}{\textbf{Seen Query, Seen Tool}} \\ \hline
GenTool  & \textbf{88.57} & \textbf{90.30} & \textbf{80.89} & \textbf{99.71} \\ \hline
Weak-to-Strong & 31.62 & 42.79 & 41.72 & 98.73 \\ \hline
Zero-to-One & 74.47 & 65.91 & 56.96 & \textbf{99.71} \\ \hline
\multicolumn{5}{c}{\textbf{Unseen Query, Unseen Tool}} \\ \hline
GenTool  & 80.58 & 77.94 & 71.18 & \textbf{100.00} \\ \hline
Weak-to-Strong & \textbf{84.38} & \textbf{80.60} & \textbf{73.23} & 99.78 \\ \hline
Zero-to-One & 74.78 & 71.62 & 67.18 & \textbf{100.00} \\ \hline
\multicolumn{5}{c}{\textbf{Unseen Query, Seen Tool}} \\ \hline
GenTool  & \textbf{91.62} & \textbf{94.50} & \textbf{88.48} & \textbf{100.00} \\ \hline
Weak-to-Strong & 73.90 & 94.48 & 87.66 & 99.73 \\ \hline
Zero-to-One & 85.71 & 87.05 & 81.59 & \textbf{100.00} \\ \hline
\end{tabular}
\caption{Detailed results of the LLaMA-3.1-8B-Instruct model trained in a single generalization simulation scenario across different scenarios}
\label{tab:detail_transfer}
\end{table*}

\section{Detailed explanation of evaluation metrics}
\label{app_eval}
\subsection{Tool Selection}
We assess the model's ability to discriminate and select the best tool by comparing the predicted tool name against the ground truth. This metric directly evaluates the model's capability to understand query contexts and match them with appropriate tools.



\subsection{Parameter Name Identification}
After selecting the appropriate tool, the model identifies the parameters required for the tool invocation based on the query. The output consists of the parameter names, and we compare the model's output parameter names with the standard parameter names, focusing on both completeness and correctness.

\subsection{Parameter Value Matching}
After selecting the tool and the required parameters, the model parses the query to fill in the content needed for the parameters. The parameter value evaluation focuses on the model's ability to extract and generate appropriate values from queries. Following \citet{huang-etal-2024-planning-creation}, we employ a normalized Levenshtein distance-based scoring function. Given the model's key-value format response $(p_k, p_v)$ and the reference answer $(y_k, y_v)$, where keys $p_k$ and $y_k$ represent the steps and values $p_v$ and $y_v$ represent task-specific results, the score is defined as follows::

\[
S = 
\begin{cases} 
F(p_v, y_v) & \text{if } p_k = y_k \\
0 & \text{if } p_k \neq y_k 
\end{cases}
\]
where \( S \) is the calculated score, and \( F \) represents the calculation function using normalized Levenshtein distance.

\subsection{Format Correctness}
Format correctness evaluates whether the model's tool invocation output is free from syntax or matching errors and can be correctly parsed into the standard JSON format. This metric ensures the generated tool calls are executable.

\section{Additional analysis}
\subsection{Data Scaling Effects} 

\begin{table*}[!t]
\definecolor{specialblue}{RGB}{29, 66, 168}
\definecolor{specialgreen}{RGB}{27, 143, 34}
\small
\centering
\begin{tabular}{l|c|c|c|c}
\hline
\multirow{2}{*}{\textbf{Method}} & \multirow{2}{*}{\textbf{Tool Selection}} & \multicolumn{2}{c|}{\textbf{Parameter}} & \multirow{2}{*}{\textbf{Format Accuracy}} \\ \cline{3-4}&& \textbf{Name}& \textbf{Value}       &\\ \hline
gpt-3.5 & 31.75 {\color{specialblue}(-16.86)} & 38.20 {\color{specialblue}(-32.14)} & 31.99 {\color{specialblue}(-27.26)} & 99.2 {\color{specialblue}(-0.09)}\\ \hline
gpt-4o mini & 54.63 {\color{specialblue}(-4.17)} & \textbf{62.53} {\color{specialblue}(-16.33)} & \textbf{55.32} {\color{specialblue}(-13.73)} & \textbf{100.0} {\color{specialgreen}(+0.07)}\\ \hline
gpt-4o& \textbf{68.96} {\color{specialblue}(-6.91)} & 59.93 {\color{specialblue}(-10.07)} & {\color{specialblue}53.31 (-10.78)} & 99.96 {\color{specialgreen}(+0.16)} \\ \hline
\end{tabular}
\caption{Results for GPT models with one demonstration that are directly relevant to the test examples.  The {\color{specialgreen}{green}} and {\color{specialblue}{blue}} values represent the differences compared to the model's zero-shot performance, respectively.}
\label{tab:incontext_tool}
\end{table*}
\begin{figure}[!tpb]
    \centering
    \includegraphics[width=1\linewidth]{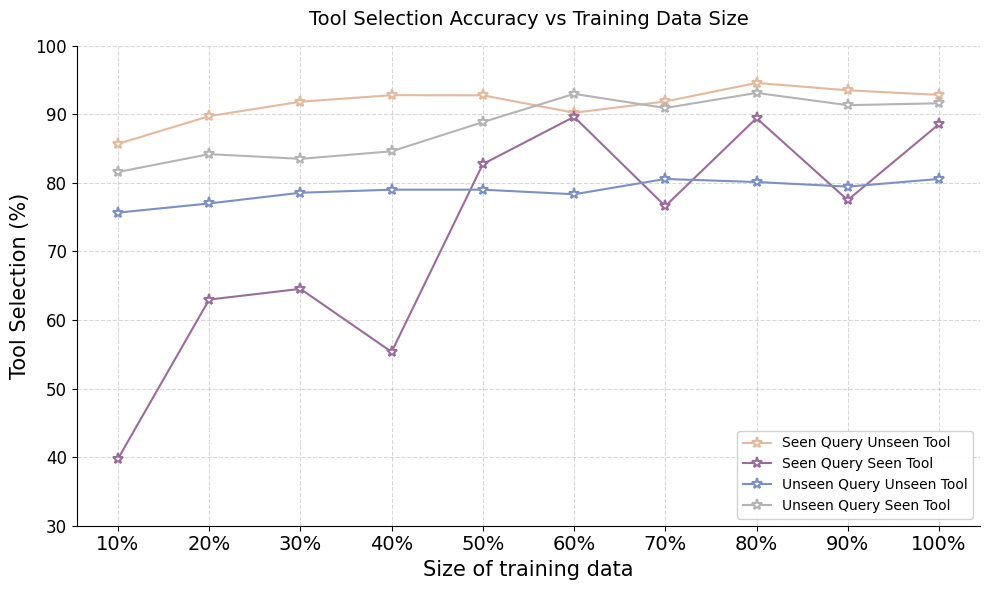}
    \caption{ Effects of different proportions of training data on various test scenarios.
   }
    \label{data_scaling}
\end{figure}

We analyze model performance across different data percentages (10\% to 100\%). Figure \ref{data_scaling} shows that performance improves significantly between 10-20\% and 40-50\% of data. Moreover, performance generally improves with more data, with varying gains across scenarios. The \textit{seen\_query\_seen\_tool} scenario benefits most from additional data, while the growth in \textit{unseen\_query\_unseen\_tool} slows down as the training data increases. Additionally, models achieve 78.61\% tool selection accuracy with just 30\% of data, comparable to the Half-Sample baseline, suggesting GenTool's effectiveness comes from structured pair construction rather than data quantity.

\subsection{Fine-grained Analysis for Transfer Learning}
We evaluate the LLaMA-3.1-8B-Instruct model fine-tuned solely on weak-to-Strong pairs and zero-to-One pairs, respectively, and analyze their performance on the test set across four scenarios. As shown in Table~\ref{tab:detail_transfer}, the Weak-to-Strong model performs better on \textit{seen\_query\_unseen\_tool} and \textit{unseen\_query\_unseen\_tool} scenarios. In contrast, the Zero-to-One model achieves better results on \textit{seen\_query\_seen\_tool} and \textit{unseen\_query\_seen\_tool} scenarios.

\begin{table}[h!]
\footnotesize
\centering
\begin{tabular}{l|ccc}
\hline
\multirow{2}{*}{\textbf{Method}} & \multirow{2}{*}{\textbf{Tool Selection}} & \multicolumn{2}{c}{\textbf{Parameter}}  \\ \cline{3-4}&& \textbf{Name}& \textbf{Value}\\ \hline
gpt-3.5 (1) & 48.61 & 70.34 & 59.25  \\ 
gpt-3.5 (2) & 49.65 & 52.55 & 44.61  \\ k
gpt-3.5 (3) & \textbf{57.89} & \textbf{72.30} & \textbf{60.24} \\ \hline

gpt-4o mini (1) & 58.80 & 78.86 & \textbf{69.05} \\ 
gpt-4o mini (2) & \textbf{76.18} & 76.05 & 66.00   \\ 
gpt-4o mini (3) & 66.88 & \textbf{79.20} & 67.97  \\ \hline

gpt-4o (1) & 75.87 & 72.00 & 64.09 \\ 
gpt-4o (2) & 79.24 & 66.68 & 59.47 \\ 
gpt-4o (3) & \textbf{84.27} & \textbf{77.14} & \textbf{67.48}  \\ \hline
\end{tabular}
\caption{Results for the GenTool  method when applied to in-context learning. (1) represents results under the zero-shot setting, (2) represents results where half of the data from zero-to-one and weak-to-one pairs is randomly sampled for demonstrations, and (3) represents results where zero-to-one and weak-to-strong pairs are used for demonstrations.}
\label{tab:in_context}
\end{table}
\subsection{In-Context Learning Capability}  
We evaluate GenTool's in-context learning using two demonstration pairs per test example (\textit{Zero-to-One} and \textit{Weak-to-Strong}), focusing on GPT-series models due to input length constraints. Table \ref{tab:in_context} shows consistent improvements with GenTool. GPT-4 achieves the highest \textit{Tool Name Selection} accuracy (84.27\%), while GPT3.5 and GPT-4o outperform the half-data baseline. Although GPT4o-mini lags in \textit{Tool Name Selection}, it surpasses the baseline in \textit{Param Name Matching} and \textit{Param Value Matching} by 3.15\% and 1.97\%, respectively, showcasing GenTool's effectiveness.

\subsection{In-context Learning for Related Demonstrations}
We further investigate the performance of GPT-series models when using training data constructed from test example-related pairs as demonstrations for in-context learning. As shown in Table~\ref{tab:incontext_tool}, all models exhibit varying degrees of performance degradation across nearly all metrics. Notably, GPT-3.5 shows the most significant drop, consistent with findings in Section~\ref{analysis_train_only}. This highlights that current large models, whether fine-tuned or used in in-context learning, fail to genuinely acquire tool-use capabilities. Instead, they primarily rely on memorizing previously seen knowledge.

\section{Case Study}
Table~\ref{tab:case1}, \ref{tab:case2} showcase two examples to demonstrate the effectiveness of GenTool . 
\begin{itemize}
    \item In the first example, the GenTool -finetuned LLaMA-8B model successfully outputs the correct priority ranking for tools and provides accurate information for tool invocation. In contrast, the zero-shot result shows that the model fails to select the correct tool.
    \item In the second example, the GenTool -finetuned LLaMA-8B model selects the most useful tool from two viable options. However, the LLaMA-8B zero-shot result fails to achieve this and instead outputs invalid tool.
\end{itemize}
These examples illustrate the superiority of GenTool  in guiding the model to make better decisions in tool selection and usage.

\section{Error Analysis}
We analyze incorrect examples where LLaMA-8B fails to generate the correct tool call. The errors can be summarized into seven categories, with examples shown in from Table \ref{tab:complete-api-docs-with-returns} to Table \ref{tab:calendar-event-api-docs} :
\begin{enumerate}
    \item \textbf{Insufficient Query Understanding:} Table \ref{tab:complete-api-docs-with-returns} shows that the model fails to parse the query correctly, resulting in incorrect parameter inputs.
    \item \textbf{Parameter Misinterpretation:} Table \ref{tab:reminder-api-docs} shows that the model provides parameter values inconsistent with tool specifications.
    \item \textbf{Query Ambiguity:} Table \ref{tab:real-estate-api-docs} shows that query ambiguity leads to hallucinated information in the tool call.
    \item \textbf{Tool Repetition:} Table \ref{tab:bond-search-api-docs} shows that the model redundantly calls the same tool multiple times, despite needing only one call.
    \item \textbf{The Second Best Tool Selection:} Table \ref{tab:job-search-api-docs} shows that the model selects the second-best tool instead of the optimal one.
    \item \textbf{Default Response Generation:} The model defaults to Generate\_Response despite available specialized tools (Table \ref{tab:calendar-event-api-docs}).
\end{enumerate}
\begin{table*}[htbp]
\begin{tabular}{p{0.15\textwidth}|p{0.8\textwidth}}
\hline
\textbf{Toolsets} & 1. flight\_pickup\_service \\
& Description: Service to arrange pick-up for flights \\
& Required parameters: \{"flight\_number": "Flight number", "arrival\_time": "Arrival time", "pickup\_location": "Pick-up location", "destination": "Destination"\} \\
& Returns: \{"status": "Pick-up service status", "service\_details": "Service details"\} \\
\cline{2-2}
& 2. car\_transfer\_service \\
& Description: Tool to arrange car transfer services \\
& Required parameters: \{"pickup\_time": "Pickup time", "pickup\_location": "Pickup location", "destination": "Destination", "vehicle\_type": "Vehicle type", "passenger\_name": "Passenger name"\} \\
& Returns: \{"transfer\_status": "Transfer status", "service\_details": "Service details", "confirmation\_code": "Confirmation code"\} \\
\cline{2-2}
& 3. simple\_car\_rental \\
& Description: Basic Car Rental Service \\
& Required parameters: \{"pickupLocation": "Pickup location", "dropOffLocation": "Drop-off location", "pickupTime": "Pickup time", "dropOffTime": "Drop-off time", "carCode": "Car code"\} \\
& Returns: \{"status": "Rental status", "cost": "Rental cost"\} \\
\cline{2-2}
& 4. airport\_dropoff\_service \\
& Description: Service to arrange drop-off to the airport \\
& Required parameters: \{"dropoff\_time": "Drop-off time", "flight\_number": "Flight number", "pickup\_location": "Pickup location", "airport\_name": "Airport name"\} \\
& Returns: \{"status": "Drop-off status", "service\_details": "Service details"\} \\
\cline{2-2}
& 5. taxi\_booking \\
& Description: Tool for booking a taxi service \\
& Required parameters: \{"pickup\_time": "Pickup time", "pickup\_location": "Pickup location", "destination": "Destination", "passenger\_name": "Passenger name", "luggage\_count": "Luggage count"\} \\
& Returns: \{"booking\_status": "Booking status", "taxi\_service\_information": "Service info", "confirmation\_number": "Confirmation number"\} \\
\hline
\textbf{User query}: & I need to know the exact pick-up and drop-off times for a car rental reservation with the code ABC123, including any grace periods for late returns. \\
\hline
\textbf{Ground truth tool calling}: & simple\_car\_rental   \{"carCode": "ABC123"\} \\
\hline
\textbf{Predicted tool calling (GenTool )}: & \{"The output of the first task: "["simple\_car\_rental", "generate\_response", "car\_transfer\_service", "flight\_pickup\_service", "airport\_dropoff\_service", "taxi\_booking"], "The output of the second task: ": ["simple\_car\_rental(carCode='ABC123')"]\}
 \\
\hline
\textbf{Predicted tool calling (Zero-Shot)}: & car\_transfer\_service \{"pickup\_time": "2023-03-15T12:00:00", "pickup\_location": "Airport", "destination": "Car Rental Agency", "vehicle\_type": "Sedan", "passenger\_name": "John Doe"\} \\
\hline
\end{tabular}
\caption{Two examples predicted by the Llama-3.1-8B-Instruct model, one from finetuning using the GenTool method and the other from performing zero-shot directly. Our method successfully enabled Llama-3.1-8B-Instruct to select the correct tool and rank the tools appropriately.}
\label{tab:case1}
\end{table*}

\begin{table*}[htbp]
\begin{tabular}{p{0.15\textwidth}|p{0.8\textwidth}}
\hline
\textbf{Toolsets}  & 1. appliance\_repair\_rescheduling \\
& Description: Request to adjust the appliance repair appointment time \\
& Required parameters: \{"originalRequestID": "Original repair request number", "newTime": "New appointment time"\} \\
& Returns: \{"rescheduling\_status": "Status", "newRequestID": "New request number"\} \\
\cline{2-2}
& 2. appliance\_repair\_status\_query \\
& Description: Appliance Repair Inquiry Tool \\
& Required parameters: \{"requestID": "Request number", "applianceType": "Appliance type", "applianceBrand": "Brand", "applianceModel": "Model", "contactInformation": "Contact info", "address": "Address"\} \\
& Returns: \{"repair\_status": "Inquiry result"\} \\
\cline{2-2}
& 3. appliance\_repair\_cancellation \\
& Description: Appliance Repair Cancellation Tool \\
& Required parameters: \{"requestID": "Request number", "cancellationReason": "Reason"\} \\
& Returns: \{"cancellation\_status": "Status"\} \\
\cline{2-2}
& 4. appliance\_repair\_request \\
& Description: Appliance Repair Request Tool \\
& Required parameters: \{"applianceType": "Type", "applianceBrand": "Brand", "applianceModel": "Model", "issueDescription": "Issue", "contactInformation": "Contact", "address": "Address", "time": "Service time"\} \\
& Returns: \{"request\_status": "Status", "requestID": "Request ID"\} \\
\cline{2-2}
& 5. repair\_service\_request \\
& Description: Request a repair service \\
& Required parameters: \{"device": "Device type", "make": "Make", "model\_number": "Model", "issue": "Issue", "phone": "Phone", "address": "Address", "appointment\_time": "Time"\} \\
& Returns: \{"request\_status": "Status", "service\_id": "Service ID"\} \\
\hline
\textbf{User query}: & Please help me schedule a repair service for my LG microwave, model LMV2031ST, which is not heating food. My contact information is 876-543-2109, and my address is 404 Cedar St. I am available on Friday at 1 PM. \\
\hline
\textbf{Ground truth tool call}: & repair\_service\_request  \{"device": "microwave", "make": "LG", "model\_number": "LMV2031ST", "issue": "not heating food", "phone": "876-543-2109", "address": "404 Cedar St", "appointment\_time": "Friday at 1 PM"\} \\
\hline
\textbf{Predicted tool calling (GenTool )}: & \{"The output of the first task: ": ["repair\_service\_request", "appliance\_repair\_request", "generate\_response", "appliance\_repair\_status\_query", "appliance\_repair\_cancellation", "appliance\_repair\_rescheduling"], "The output of the second task: ": ["repair\_service\_request(device='microwave', make='LG', model\_number='LMV2031ST', issue='not heating food', phone='876-543-2109', address='404 Cedar St', appointment\_time='Friday at 1 PM')"]\}
\\
\hline
\textbf{Predicted tool calling (Zero-shot)}: & appliance\_repair\_status\_query  \{"requestID": "876-543-2109", "applianceType": "LG microwave", "applianceBrand": "LG", "applianceModel": "LMV2031ST", "contactInformation": "876-543-2109", "address": "404 Cedar St."\} \\
\hline
\end{tabular}
\caption{Two examples were predicted by the Llama-3.1-8B-Instruct model: one using the GenTool method for fine-tuning and the other performing zero-shot inference directly. Our method successfully positioned the two useful tools before ``generate\_response'' and ultimately selected the correct tool.}
\label{tab:case2}
\end{table*}

\begin{table*}[htbp]
\begin{tabular}{p{0.15\textwidth}|p{0.80\textwidth}}
\hline
\textbf{Toolsets}  & 1. schedule\_repair\_service \\
& Description: Tool to schedule a repair service for home appliances \\
& Required parameters: \{"appliance": "Type and brand", "model": "Model number", "problem": "Issue description", "contact": "Contact number", "location": "Service location", "appointment\_time": "Service time"\} \\
& Returns: \{"status": "Status of the repair service request", "serviceID": "ID of the scheduled service"\} \\
\cline{2-2}
& 2. appliance\_repair\_cancellation \\
& Description: Appliance Repair Cancellation Tool \\
& Required parameters: \{"requestID": "Repair request number", "cancellationReason": "Reason for cancellation"\} \\
& Returns: \{"cancellation\_status": "Repair cancellation status"\} \\
\cline{2-2}
& 3. appliance\_repair\_status\_query \\
& Description: Appliance Repair Inquiry Tool \\
& Required parameters: \{"requestID": "Request number", "applianceType": "Type of appliance", "applianceBrand": "Brand name", "applianceModel": "Model number", "contactInformation": "Contact info", "address": "Service address"\} \\
& Returns: \{"repair\_status": "Repair inquiry result"\} \\
\cline{2-2}
& 4. appliance\_repair\_request \\
& Description: Appliance Repair Request Tool \\
& Required parameters: \{"applianceType": "Type of appliance", "applianceBrand": "Brand name", "applianceModel": "Model number", "issueDescription": "Problem details", "contactInformation": "Contact info", "address": "Service address", "time": "Service time"\} \\
& Returns: \{"request\_status": "Repair request status", "requestID": "Repair request ID"\} \\
\cline{2-2}
& 5. appliance\_repair\_rescheduling \\
& Description: Request to adjust the appliance repair appointment time \\
& Required parameters: \{"originalRequestID": "Original request number", "newTime": "New appointment time"\} \\
& Returns: \{"rescheduling\_status": "Rescheduling status", "newRequestID": "New repair request number"\} \\
\hline
\textbf{User query}: & I need to arrange a repair service for my LG washing machine, model WM3900HBA, which is making a loud noise during the spin cycle. Please schedule it for Saturday at 2 PM. My contact number is 987-654-3210, and my address is 456 Elm St. \\
\hline
\textbf{Ground truth tool calling}: & schedule\_repair\_service \{"appliance": "LG washing machine", "model": "WM3900HBA", "problem": "making a loud noise during the spin cycle", "contact": "987-654-3210", "location": "456 Elm St", "appointment\_time": "Saturday at 2 PM"\} \\
\hline 
\textbf{Predicted tool calling}: & schedule\_repair\_service  \{"appliance": "washing machine", "model": "WM3900HBA", "problem": "loud noise", "contact": "987-654-3210", "location": "456 Elm St", "time": "Saturday 2 PM"\} \\
\hline
\end{tabular}
\caption{LLaMA-3.1-8B-Instruct error case: Here the model lacks a comprehensive understanding of the query. During parsing, it overlooks key textual information. ``LG'' is a brand and should be associated with ``washing machine.'' }
\label{tab:complete-api-docs-with-returns}
\end{table*}
\begin{table*}[htbp]
\begin{tabular}{p{0.15\textwidth}|p{0.80\textwidth}}
\hline
\textbf{Toolsets}  & 1. add\_reminder \\
& Description: Add reminder event \\
& Required parameters: \{"reminder\_id": "ID of the reminder", "event\_title": "Event title", "reminder\_time": "Time of the reminder", "reminder\_location": "Location of the reminder", "reminder\_frequency": "Frequency of the reminder (days)", "reminder\_name": "Name of the reminder"\} \\
& Returns: \{"status": "Whether the reminder was successfully added"\} \\
\cline{2-2}
& 2. create\_event\_reminder \\
& Description: Create a one-time event reminder \\
& Required parameters: \{"event\_name": "Name of the event", "event\_time": "Time of the event", "event\_location": "Location of the event"\} \\
& Returns: \{"confirmation": "Confirmation of the event reminder creation"\} \\
\cline{2-2}
& 3. set\_reminder \\
& Description: Set a reminder for an event \\
& Required parameters: \{"event\_id": "ID of the event", "reminder\_time": "Time before the event to set the reminder"\} \\
& Returns: \{"confirmation": "Confirmation of the reminder being set"\} \\
\cline{2-2}
& 4. add\_event\_to\_calendar \\
& Description: Add an event to the calendar \\
& Required parameters: \{"event\_name": "Name of the event", "date": "Date of the event", "start\_time": "Start time of the event", "end\_time": "End time of the event", "location": "Location of the event"\} \\
& Returns: \{"status": "Status of the event addition"\} \\
\cline{2-2}
& 5. add\_event \\
& Description: Add a single event to the schedule \\
& Required parameters: \{"event\_name": "Name of the event", "start\_time": "Start time of the event", "end\_time": "End time of the event", "location": "Location of the event", "reminder": "Reminder setting (e.g., '15 minutes before')"\} \\
& Returns: \{"status": "Status of the event addition"\} \\
\hline
\textbf{User query}: & Could you create a one-time event reminder for a dinner reservation at "The Gourmet Bistro" at 7:00 PM on October 25th, 2023? \\
\hline
\textbf{Ground truth tool calling}: & create\_event\_reminder  \{"event\_name": "Dinner Reservation at The Gourmet Bistro", "event\_time": "2023-10-25T19:00:00", "event\_location": "The Gourmet Bistro"\} \\
\hline
\textbf{Predicted tool calling}:  & create\_event\_reminder  \{"event\_name": "The Gourmet Bistro", "event\_time": "2023-10-25T19:00:00", "event\_location": "Dinner Reservation"\} \\
\hline

\end{tabular}
\caption{LLaMA-3.1-8B-Instruct error case: Parameter type confusion - model places location ``gourmet bistro'' in event name field instead of location field.}
\label{tab:reminder-api-docs}
\end{table*}
\begin{table*}[htbp]
\begin{tabular}{p{0.15\textwidth}|p{0.80\textwidth}}
\hline
\textbf{Toolsets}  & 1. real\_estate\_search\_tool \\
& Description: Real estate search tool that allows for setting multiple search criteria for filtering \\
& parameters: \{"location": "Geographic location of the property", "priceRange": "Price range (format: 'min-max')", "areaRange": "Area range (format: 'min-max')", "propertyType": ["apartment", "villa", "townhouse", "condo"], "bedrooms": "Number of bedrooms", "bathrooms": "Number of bathrooms"\} \\
& Returns: \{"listings": [Array of properties with details including id, location, price, area, type, bedrooms, bathrooms, imageURL]\} \\
\cline{2-2}
& 2. market\_price\_checker \\
& Description: Market Price Checker Tool \\
& Required parameters: \{"marketType": ["Energy", "Agriculture", "Metals"], "commodity": "Commodity Name"\} \\
& Returns: \{"current\_price": "Current Market Price"\} \\
\cline{2-2}
& 3. price\_comparison\_tool \\
& Description: Product price query and comparison tool \\
& Required parameters: \{"productID1", "productID2", "date1", "date2", "price1", "price2", "storeName1", "storeName2"\} \\
& Returns: \{"price": "Product price", "lowestStore": "Store with lowest price"\} \\
\cline{2-2}
& 4. simple\_apartment\_comparator \\
& Description: Tool to compare a specific apartment with the market price of similar properties \\
& Required parameters: \{"apartmentID": "ID of apartment to compare", "marketPrice": "Market price of similar properties"\} \\
& Returns: \{"comparison": "Comparison result"\} \\
\cline{2-2}
& 5. compare\_prices \\
& Description: Tool to compare prices of two products \\
& Required parameters: \{"productId1": "First Product ID", "productId2": "Second Product ID"\} \\
& Returns: \{"priceDifference": "Price difference between products"\} \\
\hline
\textbf{User query}: & Please compare for me the finely decorated two-bedroom apartments around 100 square meters in the center of Beijing, priced between 3 million and 5 million RMB, especially the one priced at 3.2 million near the subway station, with the market price of other similar properties, so that I can make a purchase decision. \\
\hline
\textbf{Ground truth tool calling}: & real\_estate\_search\_tool  \{"location": "center of Beijing", "priceRange": "3.2 million", "areaRange": "around 100 square meters", "propertyType": "apartment", "bedrooms": 2, "bathrooms": 1"\} \{"location": "center of Beijing", "priceRange": "3-5 million", "areaRange": "around 100 square meters", "propertyType": "apartment", "bedrooms": 2, "bathrooms": 1"\} \\
\hline
\textbf{Predicted tool calling}: & simple\_apartment\_comparator  \{"apartmentID": "A123", "marketPrice": 3200000\} \\
\hline
\end{tabular}
\caption{LLaMA-3.1-8B-Instruct error case: Model incorrectly invokes simple\_apartment\_comparator and hallucinates A123, failing to recognize the query's intent for comparative property search}
\label{tab:real-estate-api-docs}
\end{table*}
\begin{table*}[htbp]
\begin{tabular}{p{0.15\textwidth}|p{0.80\textwidth}}
\hline
\textbf{Toolsets}  & 1. savings\_bond\_lookup \\
& Description: Lookup tool for savings bonds \\
& Required parameters: \{"bond\_term": "Term of the bond in months", "issue\_date": "Date when the bond was issued", "interest\_method": "Method of paying interest", "coupon\_rate\_type": "Type of coupon rate", "term\_variability": "Indicates if the term is variable", "keyword": "Keyword for bond search"\} \\
& Returns: \{"bond\_info": "Information about the bond"\} \\
\cline{2-2}
& 2. job\_info\_search \\
& Description: Job Information Query Tool \\
& Required parameters: \{"jobTitle": "Job Title", "industry": "Industry Type", "location": "Work Location", "postDate": "Date of Job Posting"\} \\
& Returns: \{"job\_descriptions": "Job Descriptions Summary", "companyName": "Company Name", "jobDescription": "Job Requirements", "salaryRange": "Salary Range"\} \\
\cline{2-2}
& 3. train\_ticket\_query \\
& Description: Train Ticket Booking Inquiry Tool \\
& Required parameters: \{"departureDate": "Departure Time", "startStation": "Departure Station", "endStation": "Arrival Station", "trainID": "Train Number", "seatType": "Seat Type"\} \\
& Returns: \{"ticket\_status": "Booking List", "order\_details": "Order Details"\} \\
\cline{2-2}
& 4. company\_credit\_check \\
& Description: Enterprise Credit Inquiry Tool \\
& Required parameters: \{"companyName": "Company name", "country": "Company location", "checkDate": "Inquiry date", "checkType": ["financial", "legal", "business"]\} \\
& Returns: \{"credit\_score": "Enterprise credit score", "credit\_rating": "Credit rating", "credit\_status": "Credit status"\} \\
\cline{2-2}
& 5. basic\_job\_search\_tool \\
& Description: A simple tool for searching job positions based on basic criteria \\
& Required parameters: \{"jobTitle": "Title", "location": "Location", "salary": "Salary range"\} \\
& Returns: \{"jobListings": [Array of jobs with company, title, location, salary]\} \\
\hline
\textbf{User query}: & I need to find the savings government bonds issued from today with a 36-month term, periodic interest payments, fixed coupon rate, and a variable term type, and view the details of the first record, especially using 'government bonds' as the keyword to obtain relevant information. \\
\hline
\textbf{Ground truth tool calling}: & savings\_bond\_lookup  \{"bond\_term": 36, "issue\_date": "today's date", "interest\_method": "periodic", "coupon\_rate\_type": "fixed", "term\_variability": "variable", "keyword": "government bonds"\} \\
\hline
\textbf{Predicted tool calling}: & savings\_bond\_lookup  \{"bond\_term": 36, "issue\_date": "today's date", "interest\_method": "periodic interest payments", "coupon\_rate\_type": "fixed coupon rate", "term\_variability": "variable term type", "keyword": "government bonds"\} savings\_bond\_lookup  \{"bond\_term": 36, "issue\_date": "today's date", "interest\_method": "periodic interest payments", "coupon\_rate\_type": "fixed coupon rate", "term\_variability": "variable term type", "keyword": "government bonds"\}\\
\hline
\end{tabular}
\caption{LLaMA-3.1-8B-Instruct error case: Redundant duplicate tool calls in response output.}
\label{tab:bond-search-api-docs}
\end{table*}

\begin{table*}[htbp]
\begin{tabular}{p{0.15\textwidth}|p{0.80\textwidth}}
\hline
\textbf{Toolsets}  & 1. part\_time\_job\_tool \\
& Description: Part-time job information collection and management tool, only applicable to college students \\
& Required parameters: \{"jobType": ["Education", "Sales", "Media", "Service Industry", "IT"], "hourlyWage": "number >= 1", "workingHours": "format 'HH:MM-HH:MM'", "jobLocation": "string", "experienceRequired": "boolean", "studentIdentityRequired": "boolean", "jobDescription": "string", "contact": "string", "postDate": "date"\} \\
& Returns: \{"success": "boolean", "jobId": "string"\} \\
\cline{2-2}
& 2. job\_info\_search \\
& Description: Job Information Query Tool \\
& Required parameters: \{"jobTitle": "Job Title", "industry": "Industry Type", "location": "Work Location", "postDate": "Date of Job Posting"\} \\
& Returns: \{"job\_descriptions": "Summary", "companyName": "Company", "jobDescription": "Requirements", "salaryRange": "Salary"\} \\
\cline{2-2}
& 3. basic\_job\_search\_tool \\
& Description: A simple tool for searching job positions based on basic criteria \\
& Required parameters: \{"jobTitle": "Title", "location": "Location", "salary": "Salary range"\} \\
& Returns: \{"jobListings": [Array of jobs with company, title, location, salary]\} \\
\cline{2-2}
& 4. job\_hiring\_tool \\
& Description: Job hiring tool for posting vacancies and filtering candidates \\
& Required parameters: \{"jobPosition": "string", "educationRequirement": ["Doctorate", "Master's", "Bachelor's", "Associate", "Other"], "workExperience": "string", "salaryRange": "string", "jobLocation": "string", "companySize": "string", "requiredSkills": "comma-separated skills"\} \\
& Returns: \{"candidates": [Array of candidates], "jobPosted": "boolean"\} \\
\cline{2-2}
& 5. savings\_bond\_lookup \\
& Description: Lookup tool for savings bonds \\
& Required parameters: \{"bond\_term": "Term of the bond in months", "issue\_date": "Date when the bond was issued", "interest\_method": "Method of paying interest", "coupon\_rate\_type": "Type of coupon rate", "term\_variability": "Indicates if the term is variable", "keyword": "Keyword for bond search"\} \\
& Returns: \{"bond\_info": "Information about the bond"\} \\
\cline{2-2}
\hline
\textbf{User query}: & Can you find job listings for a Content Writer in Sydney with a salary range of 50k-70k? \\
\hline
\textbf{Ground truth tool calling}: & basic\_job\_search\_tools \{"jobTitle": "Content Writer", "location": "Sydney", "salary": "50k-70k"\} \\
\hline
\textbf{Predicted tool calling}: & job\_info\_search  \{"jobTitle": "Content Writer", "location": "Sydney", "salaryRange": "50k-70k"\} \\
\hline
\end{tabular}
\caption{LLaMA-3.1-8B-Instruct error case: For this simple query, the model should have called the ``basic\_job\_search\_tool''. However, it incorrectly chose the ``job\_info\_search'' tool, and the parameters it used for calling this tool were inconsistent with the tool's descriptions.}
\label{tab:job-search-api-docs}
\end{table*}

\begin{table*}[htbp]
\begin{tabular}{p{0.15\textwidth}|p{0.80\textwidth}}
\hline
\textbf{Toolsets}  & 1. add\_event\_to\_calendar \\
& Description: Add an event to the calendar \\
& Required parameters: \{"event\_name": "Event name", "date": "Event date", "start\_time": "Start time", "end\_time": "End time", "location": "Event location"\} \\
& Returns: \{"status": "Event addition status"\} \\
\cline{2-2}
& 2. create\_event \\
& Description: Create a new event in the calendar \\
& Required parameters: \{"event\_title": "Event title", "date": "Event date", "start\_time": "Start time", "end\_time": "End time", "venue": "Event venue"\} \\
& Returns: \{"status": "Event creation status"\} \\
\cline{2-2}
& 3. create\_event\_label \\
& Description: Create a label for an event \\
& Required parameters: \{"event\_id": "Event ID", "event\_name": "Event name", "label\_color": "Label color", "label\_description": "Label description"\} \\
& Returns: \{"label\_creation\_status": "Label creation status"\} \\
\cline{2-2}
& 4. add\_event \\
& Description: Add a new event to the calendar \\
& Required parameters: \{"event\_name": "Event name", "event\_date": "Event date", "event\_time": "Event time"\} \\
& Returns: \{"status": "Event addition status"\} \\
\cline{2-2}
& 5. delete\_event \\ & 
Description: Delete an event from the calendar \\ &
Required parameters: 
\{ "event\_id": "Event ID", "event\_name": "Event name" \} \\
& Returns: 
\{ "status": "Event deletion status" \} \\ 
\hline
\textbf{User query}: & Please set a one-time reminder for me to attend the 'Team Building Activity' held in the company meeting room at 2 PM three days from now, and set a recurring reminder to check the project progress every Monday morning at 8 AM in my office. \\
\hline
\textbf{Ground truth tool calling}: & create\_event  \{"event\_title": "Team Building Activity", "date": "2023-10-14", "start\_time": "14:00", "end\_time": "16:00", "venue": "company meeting room"\}  create\_event \{"event\_title": "Check Project Progress", "date": "2023-10-16", "start\_time": "08:00", "end\_time": "09:00", "venue": "office"\} \\
\hline
\textbf{Predicted tool calling}: & generate\_response() \\
\hline
\end{tabular}
\caption{LLaMA-3.1-8B-Instruct error case: Model defaults to generate\_response despite availability of suitable create\_event tool.}
\label{tab:calendar-event-api-docs}
\end{table*}

\end{document}